\documentclass{l4dc2026}

\usepackage{algcompatible}

\title[Active Constraint Learning]{Active Constraint Learning in High Dimensions from Demonstrations}

\usepackage{times}
\usepackage{float}

\usepackage{booktabs}
\usepackage{adjustbox}
\usepackage{pgfplots}
\pgfplotsset{compat=1.18}
\usepackage{wrapfig}
\usepackage[OT1]{fontenc} 
\usepackage{lipsum}
\usepackage{amssymb}
\usepackage{amsmath}
\usepackage{siunitx}
\usepackage{mathtools}
\usepackage{hyperref}
\usepackage{multirow}
\usepackage{caption}
\usepackage{subcaption}

\usepackage{ifthen}
\usepackage{dsfont}
\usepackage{balance}
\usepackage{dirtytalk}
\usepackage{afterpage}
\usepackage{bm}
\usepackage{bbm}

\allowdisplaybreaks

\newcommand{\Avoid}{\mathcal{A}}

\newcommand{\D}{\mathcal{D}}

\newcommand{\E}{\mathbb{E}}

\newcommand{\Hyper}{\mathcal{H}}

\newcommand{\Line}{\mathcal{L}}

\newcommand{\MI}{\text{MI}}

\newcommand{\N}{\mathbb{N}}
\newcommand{\OR}{\textbf{OR}}

\newcommand{\ra}{\rightarrow}

\newcommand{\R}{\mathbb{R}}

\newtheorem{problem}{Problem}

\newcommand{\urk}{{\urcorner k}}

\newcommand{\eq}{\textrm{eq}}
\newcommand{\ineq}{\textrm{ineq}}

\newcommand{\KKT}{\text{KKT}}

\newcommand{\boldlambda}{\boldsymbol{\lambda}}
\newcommand{\boldnu}{\boldsymbol{\nu}}

\newcommand{\boldX}{\textbf{X}}
\newcommand{\boldY}{\textbf{Y}}

\newcommand{\iid}{\text{i.i.d.}}

\newcommand{\boldf}{\textbf{f}}
\newcommand{\boldg}{\textbf{g}}
\newcommand{\boldh}{\textbf{h}}

\newcommand{\st}{\text{ s.t. }}
\newcommand{\paren}[1]{{({#1})}}

\newcommand{\GP}{\mathcal{G}\mathcal{P}}
\newcommand{\cov}{\text{cov}}
\newcommand{\sep}{\text{sep}}
\newcommand{\boldzero}{\textbf{0}}
\newcommand{\boldone}{\textbf{1}}
\newcommand{\bolds}{\textbf{s}}
\newcommand{\boldt}{\textbf{t}}
\newcommand{\find}{\text{find}}
\newcommand{\tight}{\text{tight}}

\newcommand{\iters}{\text{iters}}
\newcommand{\ours}{\text{ours}}
\newcommand{\BL}{\text{BL}}

\author{%
 \Name{Zheng Qiu} \Email{zqiu67@gatech.edu}\\
 \addr School of Mechanical Engineering, Georgia Institute of Technology
 \AND
 \Name{Chih-Yuan Chiu} \Email{cyc@gatech.edu}\\
 \addr School of Electrical and Computer Engineering, Georgia Institute of Technology
 \AND
 \Name{Glen Chou} \Email{chou@gatech.edu}\\
 \addr Schools of Cybersecurity \& Privacy and Aerospace Engineering, Georgia Institute of Technology
}

\begin{document}

\maketitle

\begin{abstract}

We present an iterative active constraint learning (ACL) algorithm, within the learning from demonstrations (LfD) paradigm, which intelligently 
solicits informative demonstration trajectories for inferring an unknown constraint in the demonstrator's environment. 
Our approach iteratively trains a Gaussian process (GP) on the available demonstration dataset to represent the unknown constraints, 
uses the resulting GP posterior to query start/goal states, and generates informative demonstrations which are added to the dataset. 
Across simulation and hardware experiments using high-dimensional nonlinear dynamics and unknown nonlinear constraints, our method outperforms a baseline, random-sampling based method at accurately performing constraint inference from an iteratively generated set of sparse but informative demonstrations.
\end{abstract}

\begin{keywords}
Learning from demonstration, Constraint inference, Gaussian Processes, Robot safety.
\end{keywords}

\begin{figure}[ht]
    \centering\vspace{-5pt}
    \includegraphics[width=\linewidth]{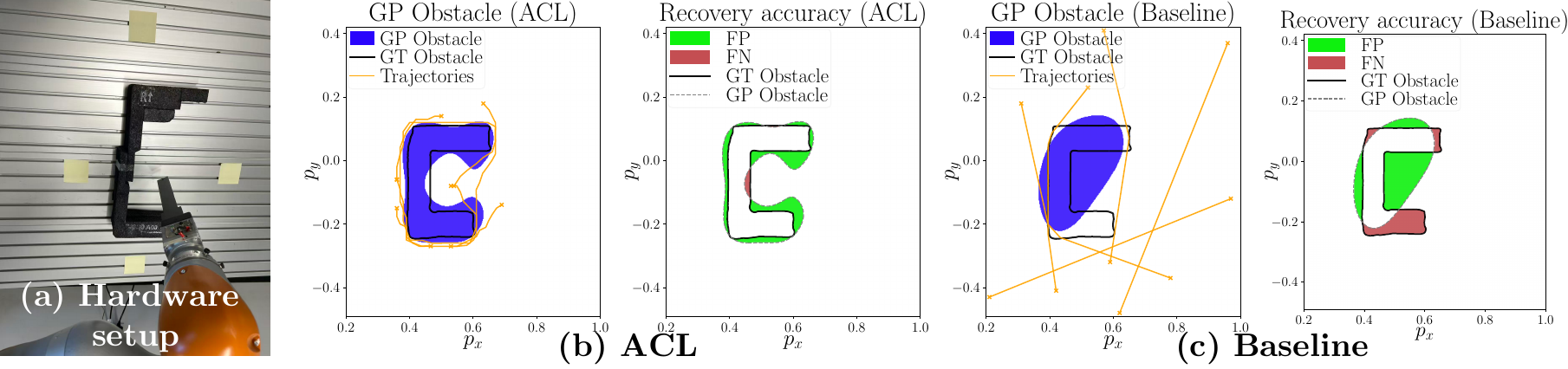}\vspace{-10pt}
    \caption{
    Given demonstrations (orange) generated via a 7-DOF robot arm,
    our GP-ACL algorithm recovers a nonlinear obstacle set with fewer instances of false positive (FP, in green) and false negative (FN, in red) errors, compared to a random-sampling baseline method. 
    }\vspace{-25pt}
    \label{fig: Front figure}
\end{figure}

\section{Introduction}
\label{sec: Introduction}

\looseness-1Recently, learning from demonstrations (LfD) via inverse optimal control (IOC) has emerged as a versatile framework for empowering robots to infer the constraints underlying an expert demonstrator's actions
\citep{Chou2022GaussianProcessConstraintLearning, Menner2021ConstrainedIOC, Chou2020LearningParametricConstraintsinHighDimensions, Armesto2017EfficientLearningofConstraintsandGenericNullSpacePolicies, PapadimitriouLi2023ConstraintInferenceinControlTasksfromExpertDemonstrationsviaInverseOptimization}, with existing methods proving effective at inferring unknown constraints from a given set of approximately locally-optimal demonstrations. However, current 
approaches implicitly assume that the demonstrations supplied for training are generated in a manner agnostic to the downstream constraint learning task, and thus the supplied data may not always be informative for reducing constraint uncertainty. As such, existing constraint inference approaches may be data-inefficient, requiring a large number of demonstrations to accurately learn unknown constraints.

To overcome this limitation, we present an iterative \textit{active} constraint learning algorithm 
built upon the Gaussian process (GP)-based constraint learning framework in \cite{Chou2022GaussianProcessConstraintLearning}, which uses the Karush-Kuhn-Tucker (KKT) optimality conditions of 
the provided demonstrations 
to train nonparametric GP-based constraint representations. 
At each iteration, our algorithm trains a Gaussian Process (GP) on a given dataset of locally-optimal demonstrations to represent the unknown constraint as a GP posterior. 
Although active learning methods have been widely explored in the context of \textit{cost} inference, to our knowledge, our work provides the first framework for active \textit{constraint} inference over continuous state spaces. Our contributions are:
\begin{enumerate}
	\item Leveraging the GP-based constraint learning framework in \cite{Chou2022GaussianProcessConstraintLearning}, 
    we sample multiple 
    GP posterior estimates
    of an unknown constraint given a demonstration dataset $\D$. The generation of multiple GP posterior samples provides a more complete description of the constraint information encoded in $\D$, as well as the remaining constraint uncertainty that cannot be resolved using $\D$.
    \vspace{-2mm}
	\item We present a GP-based active constraint learning algorithm, GP-ACL (Alg. \ref{Alg: GP-ACL}), which iteratively queries start/goal constraint states to induce the generation of informative demonstrations that reduce constraint uncertainty.
    \vspace{-2mm}
	\item We evaluate our GP-ACL algorithm by inferring high-dimensional, nonlinear constraints from the  generated demonstrations, both via 4D unicycle and 12D quadcopter dynamics in simulation, and via a 7-DOF robot arm hardware platform.
\end{enumerate}

\paragraph{Related Works}
Existing LfD-based methods have enabled constraint learning via both IOC (\cite{Chou2020LearningConstraintsfromLocallyOptimalDemonstrations, Chou2022GaussianProcessConstraintLearning, Armesto2017EfficientLearningofConstraintsandGenericNullSpacePolicies, Papadimitriou2022BayesianMethodsForConstraintInferenceInRL, Menner2021ConstrainedIOC})
and inverse reinforcement learning frameworks (\cite{McPherson2021MLConstraintInferenceFromStochasticDemonstrations, Stocking2022MaximumLikelihoodConstraintInferenceonContinuousStateSpaces, Papadimitriou2024BayesianConstraintInferencefromUserDemonstrationsBasedonMarginRespectingPreferenceModels, Singh2018RiskSensitiveInverseReinforcementLearning}).
Specifically, 
\cite{Chou2020LearningConstraintsfromLocallyOptimalDemonstrations, Chou2022GaussianProcessConstraintLearning, PapadimitriouLi2023ConstraintInferenceinControlTasksfromExpertDemonstrationsviaInverseOptimization}
use KKT optimality conditions corresponding to the provided demonstrations to formulate inverse optimization problems, from which constraint information can then be extracted.
In particular, our work builds upon and is most similar to \cite{Chou2022GaussianProcessConstraintLearning}, which likewise embeds the KKT optimality conditions for the provided demonstrations into a Gaussian process (GP) framework to represent unknown constraints. However, whereas \cite{Chou2022GaussianProcessConstraintLearning} only computes the mean and covariance functions of a trained GP posterior for constraint inference, we additionally sample GP posterior samples as surrogate functions for the unknown constraint map, to facilitate the efficient generation of informative demonstrations. Moreover, while existing constraint learning methods can enable downstream planning that is robust to epistemic uncertainty \citep{Chou2022GaussianProcessConstraintLearning, Chou2021UncertaintyAwareConstraintLearning}, they do not actively seek to reduce uncertainty through soliciting informative demonstrations. In contrast, our GP-ACL algorithm explicitly actively generates demonstrations guided by the downstream constraint learning objective, a capability not considered in prior work.

Prior work has also considered the problem of active \textit{intent} inference, with the aim of extracting the unknown reward or cost of an expert demonstrator from a provided set of trajectory demonstrations.
In particular, methods have been developed to achieve active information gathering 
\citep{Sadigh2016InformationGatheringActionsoverHumanInternalState, Sadigh2018PlanningforCarsthatCoordinatewithPeople, LiChen2025CooperativeActiveLearningBasedDualControlforExplorationandExploitationinAutonomousSearch}, intent demonstration \citep{LiBajcsy2024IntentDemonstrationinGeneralSumDynamicGames}, and uncertainty reduction \citep{Mesbah2018StochasticMPCwithActiveUncertaintyLearning, HuFisac2023ActiveUncertaintyReductionforHumanRobotInteraction}
for human-robot interaction tasks.
Meanwhile,
\cite{Akrour2012APRILActivePreferenceLearningBasedRL, Fang2017LearningHowToActivelyLearn, Lee2021PEBBLE}
devise active reward learning methods for reinforcement learning (RL). Our work likewise considers the purposeful generation of demonstrations that are maximally informative with respect to a downstream inference task. However, unlike the works listed above, our GP-ACL algorithm leverages demonstration data to recover unknown \textit{constraints}, rather than unknown intent, rewards, or costs.

Finally, our methods are related to recently developed RL-based 
(\cite{Papadimitriou2022BayesianMethodsForConstraintInferenceInRL})
and GP-based 
(\cite{LiWang2025BayesianOptimizationWithActiveConstraintLearning})
active constraint learning methods. Unlike these works, however, our method is capable of inferring unknown high-dimensional constraints in continuous and infinite state and constraint spaces and does not require both constraint-satisfying and constraint-violating behavior to guide learning, which is important for safety-critical robotics applications.

\section{Preliminaries and Problem Formulation}
\label{sec: Preliminaries and Problem Formulation}

\subsection{Demonstration Generation and KKT Optimality Conditions}
\label{subsec: Demonstration Generation and KKT Optimality Conditions}

By a \textit{demonstration}, we refer to 
a state-control trajectory $\xi := (x_1(\xi), \cdots, x_T(\xi), u_1(\xi), \cdots, u_T(\xi)) \in \R^{(n+n_i)T}$ of length $T \in \N$, where $x_t(\xi) \in \R^n$ and $u_t(\xi) \in \R^{n_i}$ respectively denote the state system and control vector identified with the demonstration $\xi$ at each time $t \in [T] := \{1, \cdots, T\}$. 
As in Sec. III-A in \cite{Chou2022GaussianProcessConstraintLearning}, we assume that each demonstration is a locally-optimal solution to 
the constrained optimization problem described below (Prob. \ref{Prob: Forward problem of demonstrator}). 
First, let $c: \R^{(n+n_i)T} \ra \R$, $\boldg_k: \R^{(n+n_i)T} \ra \R^{N_k^\ineq}$, and $\boldh_k: \R^{(n+n_i)T} \ra \R^{N_k^\eq}$ respectively encode a (possibly non-convex) cost function, as well as a set of \textit{known} inequality and equality constraints. 
In our work, we focus on smoothness-based costs of the form $c(\xi) := \sum_{t=1}^{T-1} \Vert x_{t+1} - x_t \Vert_2^2$; similar costs are often used in the constraint learning literature to encode trajectory length minimization (\cite{Chou2022GaussianProcessConstraintLearning, PapadimitriouLi2023ConstraintInferenceinControlTasksfromExpertDemonstrationsviaInverseOptimization}).
Moreover, the known equality constraint $\boldh_k(\cdot) = 0$ encodes a set of deterministic, nonlinear dynamics $x_{t+1} = f_t(x_t, u_t), \forall \ t \in [T]$, as well as constraints on the initial and final system states, which we describe in more detail in Sec. \ref{subsec: Problem Statement and Formulation}.

Next, to formulate constraints unknown to the learner, let $\phi_\sep: \R^n \ra \R^{n_c}$ denote a 
map from each system state $x$ to a \textit{constraint state} $\phi_\sep(x) \in \R^{n_c}$, at which the constraint satisfaction of the system state $x$ is then evaluated \footnote{Our formulation readily generalizes to settings in which some unknown constraints depend on control inputs.}, and let $\phi: \R^{(n+n_i)T} \ra \R^{n_cT}$ be given by $\phi(\xi) := \big(\phi_\sep(x_1(\xi)), \cdots, \phi_\sep(x_T(\xi)) \big)$. 
For each $t \in [T]$, we formulate the unknown inequality constraints below, where $g_{1, \urk}^\star, \cdots, g_{M, \urk}^\star: \R^{n_c} \ra \R$ encode scalar-valued constraints $g_{m, \urk}^\star(\cdot) \leq 0$. 
For each $t \in [T]$, we can write \say{$g_{m,\urk}^\star(\phi_\sep(x_t(\xi))) \leq 0, \forall m \in [M]$} in the more compact form of \say{$g_\urk^\star(x_t) \leq 0$}, by defining $g_\urk^\star: \R^{n_c} \ra \R$ via $g_\urk^\star(\kappa) := \max\{ g_{1, \urk}^\star(\kappa), \cdots, g_{M, \urk}^\star(\kappa) \}$ for each \textit{constraint state} $\kappa \in \R^{n_c}$. Moreover, we can stack the constraints $g_\urk^\star(\phi_\sep(x_t(\xi))) \leq 0$ across times $t \in [T]$ 
as 
$\boldg_\urk^\star(\phi(\xi)) \leq 0$, by defining $\boldg_\urk^\star: \R^{n_c T} \ra \R^T$ via $\boldg_\urk^\star(\phi(\xi)) := (g_\urk^\star(\phi_\sep(x_1(\xi))), \cdots, g_\urk^\star(\phi_\sep(x_T(\xi))) )$ for each demonstration $\xi \in \R^{(n+n_i)T}$. 

Finally, we write the demonstrator's trajectory generation problem as follows.

\begin{problem}[Demonstrator's forward problem] \label{Prob: Forward problem of demonstrator}
\begin{subequations}
\begin{align}
    \min_{\xi}. \hspace{5mm} &c(\xi) \\
    \emph{\st} \hspace{5mm} &\emph{\boldh}_k(\xi) = \emph{\boldzero}, \hspace{5mm} \emph{\boldg}_k(\xi) \leq \emph{\boldzero}, \hspace{5mm} \emph{\boldg}_\urk^\star(\phi(\xi)) \leq \emph{\boldzero},
\end{align}
\end{subequations}
\end{problem}

For each demonstration $\xi$, which forms a locally-optimal solution of Prob. \ref{Prob: Forward problem of demonstrator}, there must exist Lagrange multipliers $\boldlambda_k \in \R^{N_k^\ineq}$, $\boldlambda_\urk \in \R^T$, and $\boldnu \in \R^{N_k^\eq}$ such that the following KKT optimality conditions, denoted below by $(\xi, \boldlambda_k, \boldlambda_\urk, \boldnu) \in \KKT$, hold:
\begin{subequations} \label{Eqn: KKT, Forward problem}
\begin{align} \label{Eqn: KKT, Forward problem, Primal feasibility}
    &\boldg_\urk^\star(\phi(\xi)) \leq \boldzero, \\ \label{Eqn: KKT, Forward problem, Dual feasibility}
    &\boldlambda_k \geq \boldzero, \hspace{5mm} \boldlambda_\urk \geq \boldzero, \\ \label{Eqn: KKT, Forward problem, Complementary slackness}
    &\boldlambda_k \odot \boldg_k(\xi) = \boldzero, \hspace{5mm} \boldlambda_\urk \odot \boldg_\urk(\xi) = \boldzero, \\ \label{Eqn: KKT, Forward problem, Stationarity}
    &\nabla_\xi c(\xi) + \boldlambda_k^\top \nabla_\xi \boldg_k^\star(\xi) + \boldlambda_\urk^\top \nabla_\xi \boldg_\urk^\star(\xi) + \boldnu_k^\top \nabla_\xi \boldh_k(\xi) = \boldzero,
\end{align}
\end{subequations}
where $\odot$ denotes element-wise multiplication, and for each differentiable map $\boldf: \R^n \ra \R^m$, we define the gradient of $f$ by $[\nabla_x f]_{ij} := \frac{\partial f_i}{\partial x_j}$ for each $i \in [m]$ and $j \in [n]$. 
Above, \eqref{Eqn: KKT, Forward problem, Primal feasibility}, \eqref{Eqn: KKT, Forward problem, Dual feasibility}, \eqref{Eqn: KKT, Forward problem, Complementary slackness}, and \eqref{Eqn: KKT, Forward problem, Stationarity} encode primal and dual feasibility, complementary slackness, and stationarity conditions, respectively.
We denote the stationarity residual, i.e., the left-hand-side of \eqref{Eqn: KKT, Forward problem, Stationarity}, by $\bolds(\xi, \boldlambda_k, \boldlambda_\urk, \boldnu) \in \R^{(n+n_i)T}$, and the component of $\bolds(\xi, \boldlambda_k, \boldlambda_\urk, \boldnu)$ corresponding to partial gradients with respect to system state $x_t$ (resp., control $u_t$) by $\bolds_{x_t}(\xi, \boldlambda_k, \boldlambda_\urk, \boldnu) \in \R^n$ (resp., $\bolds_{u_t}(\xi, \boldlambda_k, \boldlambda_\urk, \boldnu) \in \R^{n_i}$).

\vspace{-8pt}
\subsection{Constraint Tightness and Constraint Gradient Extraction}
\label{subsec: Constraint Tightness and Constraint Gradient Extraction}

Suppose we are given a collection of $D$ demonstrations $S_D := \{\xi_d: d \in [D] \}$ generated by solving Prob. \ref{Prob: Forward problem of demonstrator} to local optimality, with corresponding Lagrange multipliers $\boldlambda_{d,k}$, $\boldlambda_{d,\urk}$, and $\boldnu_d$ for each $d \in [D]$.
Constraint information can be extracted from each $\xi_d$ by examining its \textit{tight} system states (if any), i.e., system states $x_t(\xi)$ at which $g_\urk^\star\big( \phi_\sep(x_t(\xi)) \big) = 0$, as well as the constraint gradient values $\nabla_{x_t} g_\urk^\star(\phi_\sep(x_t(\xi)))$ at such tight system states. Methods for both extracting tight system states and computing the corresponding gradient values were first presented in Sec. IV-A in \cite{Chou2022GaussianProcessConstraintLearning}; for completeness, we review details of these methods 
in App. \ref{subsec: App, Extraction of Tight Points and Constraint Gradient Estimates}.

\vspace{-8pt}
\subsection{Training Gaussian Process (GP)-based Constraint Representations}
\label{subsec: Training Gaussian Process (GP)-based Constraint Representations}

To learn a Gaussian Process (GP)-based representation for the unknown constraint function $g_\urk^\star(\cdot)$, we collect the constraint states and estimated constraint gradients, across the tight timesteps $t \in \boldt_\tight(\xi_d)$ of each demonstration $d \in [D]$, into the data sets $\D_\kappa$ and $\D_\nabla$, as defined below. Here, for each $d \in [D]$ and $t \in [T]$, $w_{d,t} \in \R^{1 \times n}$ denotes an estimate for the unknown constraint gradient $\nabla_{x_t} g_\urk^\star(\phi_\sep(x_t(\xi_d)))$, taken with respect to the system state $x_t$; we defer a thorough discussion of the computation of $w_{d,t}$ to Prob. \ref{Prob: Constraint Gradient Estimation} in App. \ref{subsec: App, Extraction of Tight Points and Constraint Gradient Estimates}:
\begin{align} \label{Eqn: D kappa}
    \D_\kappa &:= \big\{ \phi_\sep(x_t(\xi_d)): d \in [D], t \in \boldt_\tight(\xi_d) \big\}, \\ \label{Eqn: D nabla}
    \D_\nabla &:= \big\{ w_{d,t}: d \in [D], t \in \boldt_\tight(\xi_d) \big\}
\end{align}
Moreover, since $g_\urk^\star(\phi_\sep(x_t(\xi_d))) = 0$ for each $t \in \boldt_\tight(\xi_d)$, we also incorporate the set $\D_g := \{\boldzero \in \R^{N_\tight} \}$, where $N_\tight := \sum_{d=1}^D |\boldt_\tight(\xi_d)|$, into our training data (here, $|\cdot|$ denotes set cardinality). We then define $\boldX := \D_\kappa$ and $\boldY := (\D_g, \D_\nabla)$ to respectively be the input and output training datasets for our GP model, set $\D := (\boldX, \boldY)$, and optimize GP hyperparameters using the marginal log likelihood function (\cite{Rasmussen2006GaussianProcessesforMachineLearning}). Our GP training process parallels the approach in Sec. IV-B of \cite{Chou2022GaussianProcessConstraintLearning}.
The resulting posterior $(\tilde g|\D)(\cdot)$ of the function $g_\urk^\star$ given dataset $S_D$ is defined by the posterior mean $\E[\tilde g(\cdot) |\D ]$ and covariance $\cov[ \tilde g(\cdot)|\D]$ (see \eqref{Eqn: GP, Posterior Mean and Covariance} for details). A plausible constraint function $\hat g(\cdot)$ can be sampled from this posterior using a random Fourier feature-based approach. More details are provided in App. \ref{subsec: App, A Primer on Gaussian Processes and Their Training Process}. 

\vspace{-8pt}
\subsection{Problem Statement and Formulation}
\label{subsec: Problem Statement and Formulation}

Suppose we are given an existing demonstration dataset $S_D$ and a corresponding GP posterior $(\tilde g|\D)(\cdot)$ describing our belief of the unknown, ground truth constraint $g_\urk^\star(\cdot)$. 
We seek start ($\kappa_s$) and goal ($\kappa_g$) constraint states such that locally-optimal demonstrations $\xi$ generated while adhering to the start and goal constraints $\phi_\sep(x_1(\xi)) = \kappa_s$ and $\phi_\sep(x_T(\xi)) = \kappa_g$ maximally reduce the remaining uncertainty over the GP posterior $(\tilde g|\D)(\cdot)$.
Concretely, we formulate our problem as follows:
\begin{problem}\emph{\textbf{(Active Constraint Learning (Idealized))}} \label{Prob: Impossible Problem}
\begin{subequations} \label{Prob: Impossible Problem, Formulation}
\begin{align} \label{Prob: Impossible Problem, Objective}
    \max_{\kappa_s, \kappa_g, \xi, \boldlambda_k, \boldlambda_\urk, \boldnu}. \hspace{5mm} &\sum_{t \in \emph{\boldt}_{\emph{\tight}}(\xi)} \emph{\cov}\big[ \tilde g\big( \phi_{\emph{\sep}}(x_t(\xi)) \big)|\D \big] \\ \label{Prob: Impossible Problem, Constraints}
    \emph{\st} \hspace{5mm} &\phi_{\emph{\sep}}(x_1(\xi)) = \kappa_s, \hspace{5mm} \phi_{\emph{\sep}}(x_T(\xi)) = \kappa_g, \hspace{5mm} (\xi, \boldlambda_k, \boldlambda_\urk, \boldnu) \in \emph{\KKT}(S_D).
\end{align}
\end{subequations}
\end{problem}
Above, given a candidate demonstration $\xi$, the objective \eqref{Prob: Impossible Problem, Objective} evaluates the covariance of the GP posterior $(\tilde g|\D)(\cdot)$ at the robustly identified tight states $\{x_t(\xi): t \in \boldt_\tight(\xi) \}$ of $\xi$. Since the covariance functions of GPs serve as uncertainty measures, \eqref{Prob: Impossible Problem, Objective} captures the degree to which a candidate $\xi$ traverses regions of high constraint uncertainty with respect to the GP posterior $(\tilde g|\D)(\cdot)$. Meanwhile, \eqref{Prob: Impossible Problem, Constraints} enforce the start/goal constraints and the local optimality of $\xi$ (see Prob. \ref{Prob: Forward problem of demonstrator}).

A constraint learner who attempts to directly solve Prob. \ref{Prob: Impossible Problem} faces several challenges. First, the KKT conditions $\KKT(S_D)$ appearing in \eqref{Prob: Impossible Problem, Constraints} depend 
on knowledge of the unknown constraint $g_\urk^\star(\cdot)$ (see \eqref{Eqn: KKT, Forward problem}), which the constraint learner \textit{a priori} lacks. Moreover, for each candidate demonstration $\xi$, the tight timesteps $\boldt_\tight(\xi)$ appearing in \eqref{Prob: Impossible Problem, Objective} can only be computed by solving Prob. \ref{Prob: Tightness check} as an inner optimization loop, which renders Prob. \ref{Prob: Impossible Problem} computationally intractable. To overcome these challenges, in Sec. \ref{sec: Methods}, we decompose 
Prob. \ref{Prob: Impossible Problem} into sub-problems that allow the recovery of 
start/goal constraint states which are approximately maximally-informative for the constraint learning task.

\vspace{-8pt}
\section{Methods}
\label{sec: Methods}

Below, Sec. \ref{subsec: Optimization Routines for Approximately Solving Impossible Problem} describes optimization routines for obtaining start/goal constraint states 
which induce approximately maximally informative demonstrations,
to bypass the challenges of directly tackling Prob. \ref{Prob: Impossible Problem}, as identified in Sec. \ref{subsec: Problem Statement and Formulation}.
The prescribed optimization steps are then synthesized in Sec. \ref{subsec: GP-ACL Algorithm} into our iterative active constraint learning algorithm (Alg. \ref{Alg: GP-ACL}), which repeatedly queries start and goal states 
from which the demonstrator is likely to generate state-control trajectories which maximally reduce the remaining uncertainty over the unknown constraints.

\vspace{-8pt}
\subsection{Optimization Routines for Approximately Solving Prob. \ref{Prob: Impossible Problem}}
\label{subsec: Optimization Routines for Approximately Solving Impossible Problem}

Suppose,
as in Prob. \ref{Prob: Impossible Problem},
that we are given a demonstration set $S_D$ and corresponding GP posterior $(\tilde g|\D)(\cdot)$ and $P$ GP posterior samples $\{\hat g_p(\cdot): p \in [P] \}$. We aim to select suitable start/goal constraint states for constraint learning from each GP posterior sample $\hat g_p(\cdot)$. First, we compute the \textit{maximally informative constraint state} $\kappa_{\MI, p}$, defined as the constraint state which maximizes the covariance of the GP posterior $(\tilde g|\D)(\cdot)$ 
while remaining safe with respect to $\hat g_p(\cdot)$ 
(Prob. \ref{Prob: Computing Maximally Informative Constraint States}). 
We then approximate Prob. \ref{Prob: Impossible Problem} as the problem of searching for start/goal constraint states, using each $\kappa_{\MI,p}$ and corresponding gradient $\nabla \hat g_p(\kappa_{\MI,p})$, which induce tight demonstrations against the unknown constraint at $\kappa_{\MI,p}$, and thus provide information about 
the constraint shape near $\kappa_{\MI,p}$. 
Concretely, we present two methods for computing start/goal constraint states, as codified in Probs. \ref{Prob: Identifying a Search Direction in Hyper eta}-\ref{Prob: Orthogonal Search, Start/Goal Constraint States} and Prob. \ref{Prob: Parallel Search, Start/Goal Constraint States} below, which are tailored respectively to the settings in which the avoid set $\Avoid := \{\kappa \in \R^{n_c}: g_\urk^\star(\kappa) > 0\}$ (i) is locally-convex near $\kappa_{\MI,p}$, or (ii) is not.
Since the constraint learner lacks \textit{a priori} knowledge of the geometry of $\Avoid$, our GP-ACL approach prescribes the application of \textit{both} methods to generate two start/goal constraint state pairs.

\paragraph{Finding Maximally-Informative Constraint States $\{\kappa_{\MI,p}: p \in [P] \}$} 

We begin by computing, for each GP posterior sample $\hat g_p(\cdot)$, a maximally-informative constraint state $\kappa_{\MI,p}$ which maximizes the covariance of the GP posterior $(\tilde g|\D)(\cdot)$ while remaining tight with respect to $\hat g_p(\cdot)$.

\begin{problem}\emph{(\textbf{Computing Maximally-Informative Constraint States $\{\kappa_{\MI,p}: p \in [P] \}$})}
\label{Prob: Computing Maximally Informative Constraint States}
\begin{subequations}
\begin{align}
    \kappa_{\emph{\MI}, p} \in \arg \max_{\kappa \in \R^{n_c}}. \hspace{5mm} &\emph{\cov}[\tilde g(\kappa)|\D] \\
    \emph{\st} \hspace{5mm} &\hat g_p(\kappa) = 0.
\end{align}
\end{subequations}
\end{problem}

\begin{figure}[ht]\vspace{-5pt}
    \centering
        \includegraphics[width=0.85\linewidth]{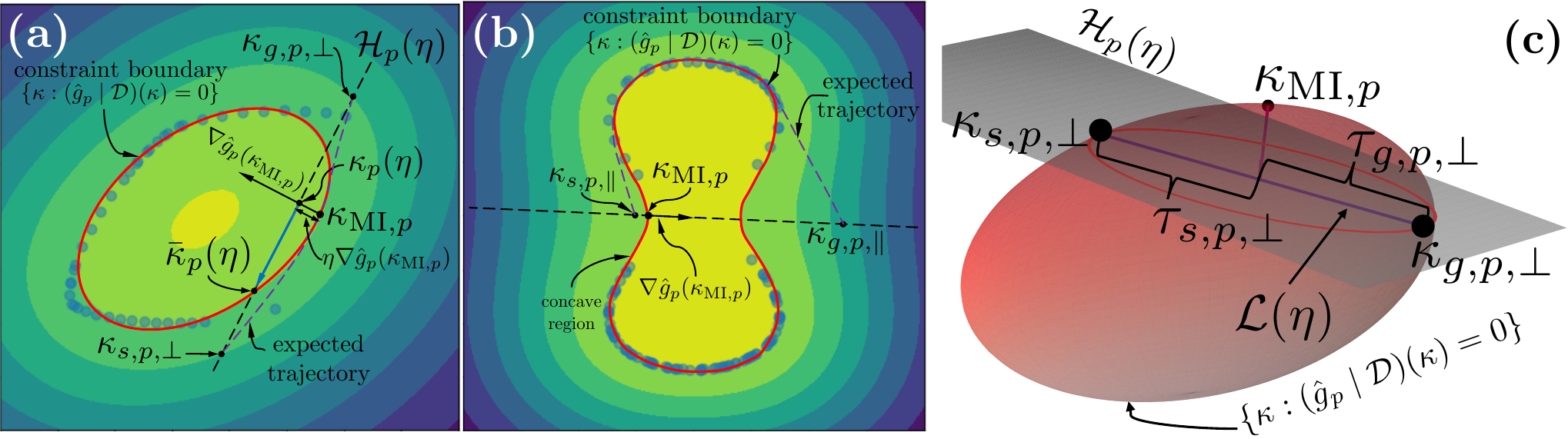}
        \vspace{-10pt}\caption{
        (a) Our method for finding start / goal constraint states 
        that induce tight demonstrations. Blue dots represent tight points collected from demonstrations and used to train the GP constraint representation (see Sec. \ref{subsec: Constraint Tightness and Constraint Gradient Extraction},  \ref{subsec: Training Gaussian Process (GP)-based Constraint Representations}, and \ref{subsec: App, Extraction of Tight Points and Constraint Gradient Estimates}). 
        To handle locally-convex constraint boundaries, we search along a hyperplane $\Hyper(\eta)$ orthogonal to the 
        gradient $\nabla \hat g_p(\kappa_{\MI,p})$. (b) To handle locally-concave constraint boundaries, we search along the gradient $\nabla \hat g_p(\kappa_{\MI,p})$. (c) Visualization of the locally-convex case in 3D.
    }
    \label{fig: Orthogonal and Parallel Search for Start/Goal Constraint States}
\end{figure}

Below, we introduce separate schemes for generating start/goal constraint states under the two distinct scenarios in which the avoid set $\Avoid$ either is locally convex near $\kappa_{\MI,p}$ (Fig. \ref{fig: Orthogonal and Parallel Search for Start/Goal Constraint States}a) or is not (Fig. \ref{fig: Orthogonal and Parallel Search for Start/Goal Constraint States}b).

\paragraph{Optimizing Start/Goal Constraint States 
in a Hyperplane $\Hyper(\eta)$ Orthogonal to $\nabla \hat g_p(\kappa_{\MI,p})$}

We begin by formulating a search method for start/goal constraints that is adapted for the setting in which the avoid set $\Avoid$ is locally-convex near $\kappa_{\MI,p}$, and thus the boundary of $\Avoid$ curves away from $\kappa_{\MI,p}$ in the direction of $\nabla \hat g_p(\kappa_{\MI, p})$ (see Fig. \ref{fig: Orthogonal and Parallel Search for Start/Goal Constraint States}a).
Since we aim to induce demonstrations that are taut against the boundary of $\Avoid$ (and thus reveal information about $\Avoid$) near $\kappa_{\MI, p}$, we wish to select start/goal constraint states from which the induced demonstration closely approximates the geometry of $\Avoid$'s boundary near $\kappa_{\MI, p}$. To this end, we first fix a constraint state $\kappa_p(\eta)$ that is slightly offset from $\kappa_{\MI,p}$ in the direction of $\nabla \hat g_p(\kappa_{\MI, p})$, with the degree of offset measured by the step size $\eta$.
The purpose of defining a constraint state $\kappa_p(\eta)$ offset from $\kappa_{\MI,p}$ is to increase the likelihood that the resulting demonstration is tight against the avoid set $\Avoid$ at $\kappa_{\MI,p}$, i.e., the true constraint $g_\urk^\star$ is active at $\kappa_{\MI,p}$, for a constraint that is locally-convex around $\kappa_{\MI,p}$ (see Fig. \ref{fig: Orthogonal and Parallel Search for Start/Goal Constraint States}a). 
We then search for start/goal constraint states $(\kappa_{s,p,\perp}, \kappa_{g,p,\perp})$ outside $\Avoid$ but within a hyperplane $\Hyper_p(\eta)$ which is orthogonal to $\nabla \hat g_p(\kappa_{\MI,p})$ and passes through $\kappa_p(\eta)$ (see Fig. \ref{fig: Orthogonal and Parallel Search for Start/Goal Constraint States}a). As illustrated in Fig. \ref{fig: Orthogonal and Parallel Search for Start/Goal Constraint States}a, the demonstration trajectory generated from $(\kappa_{s,p,\perp}, \kappa_{g,p,\perp})$ is likely to curve away from $\kappa_{\MI,p}$ in the direction of $\nabla \hat g_p(\kappa_{\MI,p})$, similar to the boundary of $\Avoid$ near $\nabla \hat g_p(\kappa_{\MI,p})$, and yield a tight point at $\kappa_{\MI,p}$ at which the true constraint $g_\urk^\star$ is active.

Concretely, we first fix a suitably small step size $\eta$, and define:
\begin{align} \label{Eqn: kappa p eta (offset)}
	\kappa_p(\eta) &:= \kappa_{\MI,p} + \eta \nabla \hat g_p(\kappa_{\MI,p}), \hspace{5mm} \forall p \in [P], \\ \label{Eqn: Hyper p eta (offset)}
	\Hyper_p(\eta) &:= \big\{ \kappa \in \R^{n_c}: \nabla \hat g_p(\kappa_{\MI,p}) (\kappa - \kappa_p(\eta) ) = 0 \ \big\}, \hspace{5mm} \forall p \in [P].
\end{align}
To fix a search direction within $\Hyper(\eta)$ starting from $\kappa_p(\eta)$, we aim to find a constraint state $\bar \kappa_p(\eta)$ in $\Hyper(\eta)$ that is of maximum possible distance from $\kappa_p(\eta)$ while remaining \textit{unsafe} with respect to the GP posterior sample $\hat g_p(\cdot)$ (Prob. \ref{Prob: Identifying a Search Direction in Hyper eta}).
Intuitively, $\bar \kappa_p(\eta)$ is a constraint state on the boundary of $\Avoid$ that lies on $\mathcal{H}_p(\eta)$ which we aim to locate, to obtain a search direction (in the form of $\kappa_p(\eta) - \bar \kappa_p(\eta)$) along which safe start/goal constraint states can be located.

\begin{problem}\emph{(\textbf{Identifying a Search Direction}} $\bar \kappa_p(\eta) - \kappa_p(\eta)$ \emph{\textbf{in}} $\Hyper(\eta)$\emph{)}
\label{Prob: Identifying a Search Direction in Hyper eta}
\begin{subequations} 
\begin{align}
    \bar \kappa_p(\eta) \in \arg \max_{\kappa \in \R^{n_c}}. \hspace{5mm} &\Vert \kappa - \kappa_p(\eta) \Vert_2 \\
    \emph{\st} \hspace{5mm} &\kappa \in \Hyper(\eta), \hspace{5mm} \hat g_p(\kappa) > 0.
\end{align}
\end{subequations}
\end{problem}

We then search for start/goal constraint states along the line $\Line(\eta) := \{\kappa_p(\eta) + t (\bar \kappa_p(\eta) - \kappa_p(\eta)): t \in \R \}$ on $\Hyper(\eta)$ (Prob. \ref{Prob: Orthogonal Search, Start/Goal Constraint States}). In effect, $\Line(\eta)$ identifies, among all vectorial directions in $\Hyper(\eta)$ starting from $\kappa_p(\eta)$, the direction along which constraint states remain unsafe for the longest distance away from $\kappa_p(\eta)$ (see Fig. \ref{fig: Orthogonal and Parallel Search for Start/Goal Constraint States}c). 
We note that, from start/goal constraint states selected on $\Line(\eta)$ outside of the avoid set $\Avoid$, the demonstrator is likely to generate trajectories that are taut against $\Avoid$, in order to minimize path length while ensuring feasibility.

Concretely, for each $p \in [P]$, we search for start/goal constraint states on $\Line(\eta)$ that are as close to $\kappa_p(\eta)$ as possible while remaining safe. To encode safety, we consider constraint states $\kappa$ that are (i) strictly safe with respect to $\hat g_p(\cdot)$ by a margin $\delta > 0$, and (ii) safe with respect to the GP posterior $(\tilde g|\D)(\cdot)$ with probability at least $\beta > 0$, where both $\delta$ and $\beta$ are algorithm design parameters.
Mathematically, the two safety conditions described above are given by:
\begin{align} \label{Eqn: Orthogonal Search, Safety Conditions}
    \textstyle\hat g_p(\kappa) \leq - \delta, \hspace{1cm} \Phi\left( \frac{\E[\tilde g(\kappa)|\D]}{\sqrt{\cov[\tilde g(\kappa)|\D]}} \right) \geq \beta,
\end{align}
where $\Phi(\cdot)$ is the cumulative distribution function (CDF) of the unit Gaussian distribution. We formulate Prob. \ref{Prob: Orthogonal Search, Start/Goal Constraint States} below to compute optimal start and goal constraint states, 
denoted respectively by $\kappa_{s,p,\perp}$ and $\kappa_{g,p,\perp}$, 
along different segments of $\Line(\eta)$ corresponding to the selection of negative or positive values of the scale parameter $\tau$
, which measures distance from the unsafe constraint state $\kappa_p(\eta)$. 
We choose $\tau^2$ as our objective in \eqref{Eqn: Objective, Orthogonal Search, Start/Goal Constraint States} to compel the start/goal constraint states to be close to 
$\kappa_p(\eta)$ while remaining probabilistically safe in the sense of \eqref{Eqn: Orthogonal Search, Safety Conditions}, to increase the likelihood of generating demonstrations that are close to, and tight against, the avoid set $\Avoid$ (see Fig. \ref{fig: Orthogonal and Parallel Search for Start/Goal Constraint States}a).

\begin{problem}\emph{(\textbf{Optimizing Start/Goal Constraint States in $\Hyper(\eta)$ Along $\bar \kappa_p(\eta) - \kappa_p(\eta)$})}
\label{Prob: Orthogonal Search, Start/Goal Constraint States}
\begin{subequations} \label{Eqn: Orthogonal Search, Start/Goal Constraint States}
\begin{align} \nonumber
    &(\tau_{s,p,\perp}, \kappa_{s,p,\perp}) \ \emph{\OR} \ (\tau_{g,p,\perp}, \kappa_{g,p,\perp}) \\ \label{Eqn: Objective, Orthogonal Search, Start/Goal Constraint States}
    \in \ &\arg 
    \min_{\substack{\tau \in \R \\ \kappa \in \R^{n_c}}} 
    \tau^2 \\ \label{Eqn: Constraints, Orthogonal Search, Start/Goal Constraint States}
    &\hspace{9mm} \emph{\st} \kappa = \kappa_p(\eta) + \tau \big(\bar \kappa_p(\eta) - \kappa_p(\eta) \big), \hspace{5mm} \kappa \emph{\text{ satisfies }} \eqref{Eqn: Orthogonal Search, Safety Conditions}, \hspace{5mm} \tau < 0 \ \emph{\OR} \ \tau > 0.
\end{align}
\end{subequations}
\end{problem}

\begin{remark}
While numerically solving Prob. \ref{Prob: Orthogonal Search, Start/Goal Constraint States}, we often encode the constraints \eqref{Eqn: Orthogonal Search, Safety Conditions} as penalties in the cost \eqref{Eqn: Objective, Orthogonal Search, Start/Goal Constraint States} with large weights, to bypass the issue that initializations of $\kappa$ may fail to satisfy \eqref{Eqn: Orthogonal Search, Safety Conditions}.
\end{remark}

\paragraph{Optimizing Start/Goal Constraint States 
Along $\nabla \hat g_p(\kappa_{\MI,p})$}

If the true constraint set were in fact \textit{not} locally convex near $\kappa_{\MI,p}$, the boundary of $\Avoid$ may curve away from $\kappa_{\MI,p}$ in the direction of $-\nabla \hat g_p(\kappa_{MI,p})$ (see Fig. \ref{fig: Orthogonal and Parallel Search for Start/Goal Constraint States}b).
In this case, start/goal constraints located by searching along $\pm \nabla \hat g_p(\kappa_{\MI, p})$  from $\kappa_{\MI,p}$, which we denote by $(\kappa_{s,p,\parallel}, \kappa_{g,p,\parallel})$, are likely to generate demonstrations that are tight against $\Avoid$, since such demonstrations may also exhibit curvature away from $\kappa_{\MI,p}$ in the direction of $-\nabla \hat g_p(\kappa_{MI,p})$, similar to the boundary of $\Avoid$.
An illustration is provided in Fig. \ref{fig: Orthogonal and Parallel Search for Start/Goal Constraint States}b, and the computation of $(\kappa_{s,p,\parallel}, \kappa_{g,p,\parallel})$ is described in detail in Prob. \ref{Prob: Parallel Search, Start/Goal Constraint States}.
We note that Probs. \ref{Prob: Computing Maximally Informative Constraint States}-\ref{Prob: Parallel Search, Start/Goal Constraint States} are  efficiently solvable via the IPOPT solver \citep{wachter2006IPOPT} in Casadi \citep{Andersson2019Casadi}.
Given that 
$\Avoid$
may or may not be locally-convex near $\kappa_{\MI,p}$, searching for start/goal constraint states 
in directions both orthogonal and parallel to the gradient $\nabla \hat g_p(\kappa_{\MI, p})$ (Probs. \ref{Prob: Identifying a Search Direction in Hyper eta}-\ref{Prob: Parallel Search, Start/Goal Constraint States})
increases the likelihood of generating demonstrations that are tight against the unknown constraint.

Concretely, Prob. \ref{Prob: Parallel Search, Start/Goal Constraint States} considers 
a variant of Prob. \ref{Prob: Orthogonal Search, Start/Goal Constraint States} which searches for start/goal states in the direction of $\nabla \hat g_p(\kappa_{\MI,p})$ from $\kappa_p(\eta)$, rather than along the direction $\bar \kappa_p(\eta) - \kappa_p(\eta)$, with the same aim of inducing tight demonstrations 
that are taut against
the constraint boundary (see Fig. \ref{fig: Orthogonal and Parallel Search for Start/Goal Constraint States}b).

\begin{problem} \emph{(\textbf{Optimizing Start/Goal Constraints Along $\nabla \hat g_p(\kappa_{\MI,p})$})}
\label{Prob: Parallel Search, Start/Goal Constraint States}
\begin{subequations}
\begin{align} \nonumber
    &(\tau_{s,p,\parallel}, \kappa_{s,p,\parallel}) \ \emph{\OR} \ (\tau_{g,p,\parallel}, \kappa_{g,p,\parallel}) \\
    \in \ &\arg 
    \min_{\substack{\tau \in \R \\ \kappa \in \R^{n_c}}} 
    \tau^2 \\
    &\hspace{9mm} \emph{\st} \kappa = \kappa_p(\eta) + \tau \nabla \hat g_p(\kappa_{\emph{\MI},p})^\top, \hspace{5mm} \kappa \emph{\text{ satisfies }} \eqref{Eqn: Orthogonal Search, Safety Conditions}, \hspace{5mm} \tau < 0 \ \emph{\OR} \ \tau > 0.
\end{align}
\end{subequations}
\end{problem}

\subsection{GP-ACL Algorithm}
\label{subsec: GP-ACL Algorithm}

We present our algorithm for active constraint learning (GP-ACL) in Alg. \ref{Alg: GP-ACL}.
Concretely, given an available demonstration dataset $S_D^\paren{i}$ at each iteration $i \in [N_\iters]$, we solve Prob. \ref{Prob: Tightness check} to extract tight points and perform GP learning via Prob. \ref{Prob: Constraint Gradient Estimation} to construct the stochastic GP posterior $(\tilde g^\paren{i}|\D)(\cdot)$ as an estimate of the unknown constraint function $g_\urk^\star(\cdot)$ 
(Lines \ref{Algline: Learn GP, 1}-\ref{Algline: Learn GP, 2}). 
From $(\tilde g^\paren{i}|\D)(\cdot)$, we compute the posterior mean $\E[\tilde g^\paren{i}(\cdot)|\D]$ and randomly draw posterior samples $\{\hat g_p^\paren{i}(\cdot): p \in [P] \}$. 
Next, by solving Prob. \ref{Prob: Identifying a Search Direction in Hyper eta}-\ref{Prob: Parallel Search, Start/Goal Constraint States}, we query start and goal constraint states that are likely to compel the demonstrator to produce constraint-revealing trajectories 
(Lines \ref{Algline: Learn Maximally informative constraint states}-\ref{Algline: Parallel method, for computing start/goal constraint states}). 
Finally, after new demonstrations are generated (see Prob. \ref{Prob: Forward problem of demonstrator}), we insert the newly generated demonstrations and their corresponding robustly identified constraint state and gradient values (see Sec. \ref{subsec: Training Gaussian Process (GP)-based Constraint Representations}) into $S_D^\paren{i}$ to form an updated dataset $S_D^\paren{i+1}$, and proceed to the next iteration of our algorithm 
(Lines \ref{Algline: Generate trajectories, 1}-\ref{Algline: Update Demonstration set}).

\begin{algorithm2e}[t!]
    \KwData{$N_\iters$, $\alpha$, $n_\ell$, $\eta$, initial demonstration set $S_D^\paren{1}$, $\delta$, $\beta$
    }
    \KwResult{$S_D^\paren{N_\iters}$
    }

\begin{algorithmic}[1]

    \FOR{$i=1, \cdots, N_\iters-1$}

        \STATE $(\tilde g^\paren{i}|\D^\paren{i})(\cdot), \{\hat g_p^\paren{i}(\cdot): p \in [P]\} \gets$ Learn GP posterior and draw GP posterior samples
        \label{Algline: Learn GP, 1}
        
        \STATE $\hspace{5mm}$ via Probs. \ref{Prob: Tightness check}-\ref{Prob: Constraint Gradient Estimation}, using the hyperparameters $\alpha$ and $n_\ell$
        \label{Algline: Learn GP, 2}
        
        \STATE $\{\kappa_{\MI,p}^\paren{i}: p \in [P] \} \gets$ Compute maximally-informative constraint states via Prob. \ref{Prob: Computing Maximally Informative Constraint States}
        \label{Algline: Learn Maximally informative constraint states}

        \STATE $\kappa_p(\eta), \Hyper(\eta) \gets$ Compute offset constraint state and hyperplane via \eqref{Eqn: kappa p eta (offset)} and \eqref{Eqn: Hyper p eta (offset)}
        \label{Algline: Compute offset constraint state and hyperplane}

        \STATE $S_{\kappa,1}^\paren{i} := \{(\kappa_{s,p,\perp}^\paren{i}, \kappa_{g,p,\perp}^\paren{i}): p \in [P] \} \gets$ Compute start/goal constraint states via Probs. \ref{Prob: Identifying a Search Direction in Hyper eta}-\ref{Prob: Orthogonal Search, Start/Goal Constraint States}
        \label{Algline: Orthogonal method, for computing start/goal constraint states}
        
        \STATE $S_{\kappa,2}^\paren{i} := \{(\kappa_{s,p,\parallel}^\paren{i}, \kappa_{g,p,\parallel}^\paren{i}): p \in [P] \} \gets$ Compute start/goal constraint states via Prob. \ref{Prob: Parallel Search, Start/Goal Constraint States}
        \label{Algline: Parallel method, for computing start/goal constraint states}

        \STATE 
        $
        \tilde S_D^\paren{i}
        \gets \forall \ (\kappa_{s,j}^\paren{i}, \kappa_{g,j}^\paren{i}) \in S_{\kappa,1}^\paren{i} \bigcup S_{\kappa,2}^\paren{i},$ solve Prob. \ref{Prob: Forward problem of demonstrator} while enforcing 
        \label{Algline: Generate trajectories, 1}
        
        \STATE $\hspace{5mm}$ $\phi_\sep(x_o(\xi_{j}^\paren{i})) = \kappa_{s,j}^\paren{i}$ and $\phi_\sep(x_T(\xi_{j}^\paren{i})) = \kappa_{g,j}^\paren{i}$
        \label{Algline: Generate trajectories, 2}
        
        \STATE $S_D^\paren{i+1} \gets S_D^\paren{i} \bigcup 
        \tilde S_D^\paren{i}
        $
        \label{Algline: Update Demonstration set}
    \ENDFOR
 \caption{Gaussian Process-based Active Constraint Learning (GP-ACL) Algorithm.}
 \label{Alg: GP-ACL}
\end{algorithmic}
\end{algorithm2e}

\vspace{-11pt}
\section{Experiments}
\label{sec: Experiments}
\vspace{-3pt}

To evaluate our GP-ACL algorithm, we perform constraint learning tasks on simulations using double integrator, 4D unicycle, 12D quadcopter, and 7-DOF robot arm dynamics, and on hardware platforms using a 7-DOF robot arm. Below, Sec. \ref{subsec: Experiment Setup} introduces parameter settings shared across our experiments, while Sec. \ref{subsec: Experiment Results} presents a select subset of our experiment results. For additional experiments, see App. \ref{sec: App, Supplementary Material for Experiments Section}.

\vspace{-10pt}
\subsection{Experiment Setup}
\label{subsec: Experiment Setup}
\vspace{-3pt}

Our experiments involve the dynamics, constraints, and costs listed below. Given a state vector $x_t$, we use $p_t \in \R^3$ and $p_{x,t}$, $p_{y,t}$, and $p_{z,t} \in \R$ to respectively denote 
the overall 3D position vector and the $x$-, $y$-, and $z$- position coordinates encoded by $x$.

\paragraph{Dynamics models}

We infer constraints from demonstrations generated using 2D and 3D double-integrator, 4D unicycle, 12D quadcopter (\cite{sabatino2015PhDThesisquadrotor}), and 7 DOF robot arm (\cite{MurraySastry1994AMathematicalIntroductiontoRoboticManipulation}) dynamics. In our experiments, we discretize the above continuous-time dynamics at intervals of $\Delta t = 1$ and set a time horizon of $T = 30$, unless stated otherwise.

\paragraph{Constraints}
We consider eight types of unknown nonlinear constraints $\{g_{\urk,i}^\star: i \in [8]\}$, each defining a corresponding obstacle set $\Avoid_i$ that demonstrations must avoid. We define the constraint space for each obstacle set to be either the configuration space of the robot arm, or in the 2D/3D Cartesian coordinate system for all other dynamics. For the mathematical definition of each constraint type, see Appendix \ref{subsec: App, Constraint Types}. 
In addition to the nonlinear constraints mentioned above, each demonstration trajectory must satisfy start/goal constraints, as described in Secs. \ref{sec: Preliminaries and Problem Formulation}-\ref{sec: Methods}.

\paragraph{Costs} All demonstrations in the following experiments are generated via the smoothness cost $c(x) := \sum_{t=1}^{T-1} \Vert p_{t+1} - p_t \Vert_2^2$, which compels each demonstration to minimize the total distance traversed between the prescribed start/goal constraints while avoiding the prescribed obstacle sets.

\paragraph{Algorithm Implementation}
When running our GP-ACL algorithm (Alg. \ref{Alg: GP-ACL}), unless otherwise specified, we set $\alpha = 0.3$ and use $n_\ell = 1000$ random Fourier basis functions to train the GP posterior for constraint representation (\eqref{Eqn: GP Posterior Sampling} and Alg. \ref{Alg: GP-ACL}, Lines \ref{Algline: Learn GP, 1}-\ref{Algline: Learn GP, 2}), with $N_\iters = 3$ and $\delta = 0.001$. We select $\beta = 0.55$ to compute the start/goal constraint states $(\kappa_{s,p,\perp}, \kappa_{g,p,\perp})$ in Alg. \ref{Alg: GP-ACL}, Line \ref{Algline: Orthogonal method, for computing start/goal constraint states}, and $\beta = 0.3$ and $\beta = 0.55$ to compute $\kappa_{s,p,\parallel}$ and $\kappa_{g,p,\parallel}$, respectively, on Line \ref{Algline: Parallel method, for computing start/goal constraint states}.

\paragraph{Metrics for Assessing Constraint Recovery Accuracy}
To evaluate the constraint recovery accuracy of our GP-ACL (resp., a random-sampling baseline) method, we randomly generate $n_s$ samples in the constraint space and report the fraction $\gamma_\ours$ (resp., $\gamma_\BL$) of sampled constraint states at which 
constraint satisfaction or violation was accurately predicted. 
For both methods, we also visualize constraint states corresponding to false positive (FP) and false negative (FN) errors, defined respectively as incidents in which safe constraint states are mistakenly marked as unsafe, and vice versa.

\begin{figure}[ht]
    \centering\vspace{-10pt}
    \includegraphics[width=\linewidth]{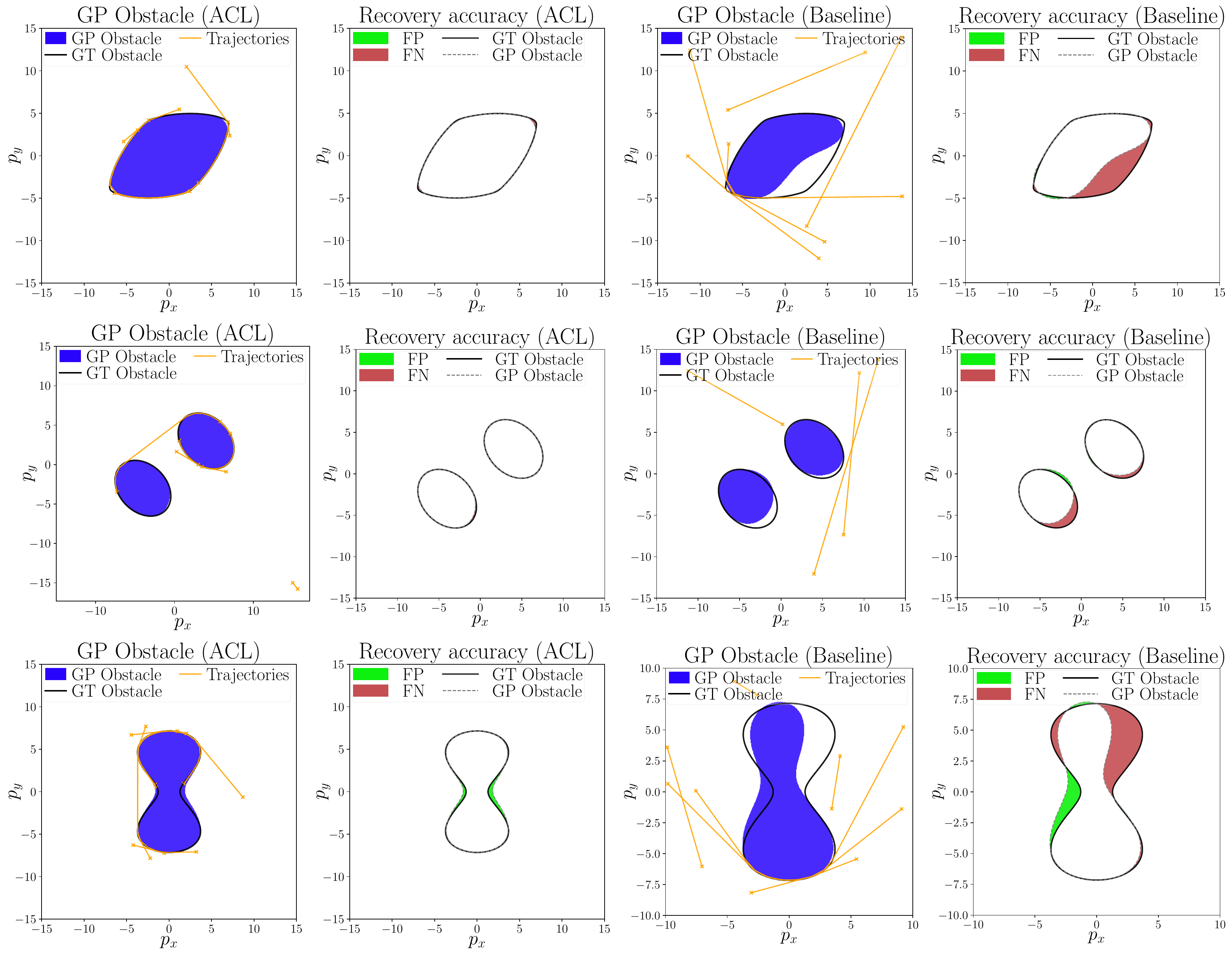}\vspace{-10pt}
    \caption{
    Our GP-ACL algorithm (left half) outperforms the random sampling baseline (right half) in accurately recovering constraints $g_{\urk,1}^\star$ (top), $g_{\urk,2}^\star$ (middle), and $g_{\urk,3}^\star$ (bottom) from unicycle dynamics demonstrations, with fewer false positive (green) and false negative (red) errors.
    }
    \label{fig: Exp_fig_3___unicycle}
\end{figure}

\vspace{-10pt}
\subsection{Experiment Results}
\label{subsec: Experiment Results}

\paragraph{Unicycle Simulations}
We evaluate our GP-ACL algorithm by recovering the three complex nonlinear constraints $g_{\urk,1}^\star$, $g_{\urk,2}^\star$, and $g_{\urk,3}^\star$, as defined in App. \ref{subsec: App, Constraint Types} and visualized in Fig. \ref{fig: Exp_fig_3___unicycle}. 
Here, we set $\alpha = 0.3$ when sampling GP posteriors. 
Across $n_s = 2500$ sampled constraint states, our GP-ACL algorithm accurately predicts the safeness or unsafeness of each sample with accuracy $\gamma_\ours = 0.9996$, $1.0$, and $0.9956$ for the constraints $g_{\urk,1}^\star$, $g_{\urk,2}^\star$ and $g_{\urk,3}^\star$, respectively,
while
the random sampling-based baseline method yielded accuracies of only $\gamma_\BL = 0.9788$, $0.9912$, and $0.9492$, respectively.
In particular, although both the baseline method and our approach accurately classified most of the space within each obstacle set, our method incurred significantly lower misclassification rates near obstacle boundaries, resulting in higher constraint representation quality (Fig. \ref{fig: Exp_fig_3___unicycle}).
Overall, our numerical results illustrate that our GP-ACL algorithm outperforms the random sampling baseline in recovering \textit{a priori} unknown constraint sets with complex boundaries.

\begin{figure}[ht]
    \centering
    \includegraphics[width=\linewidth]{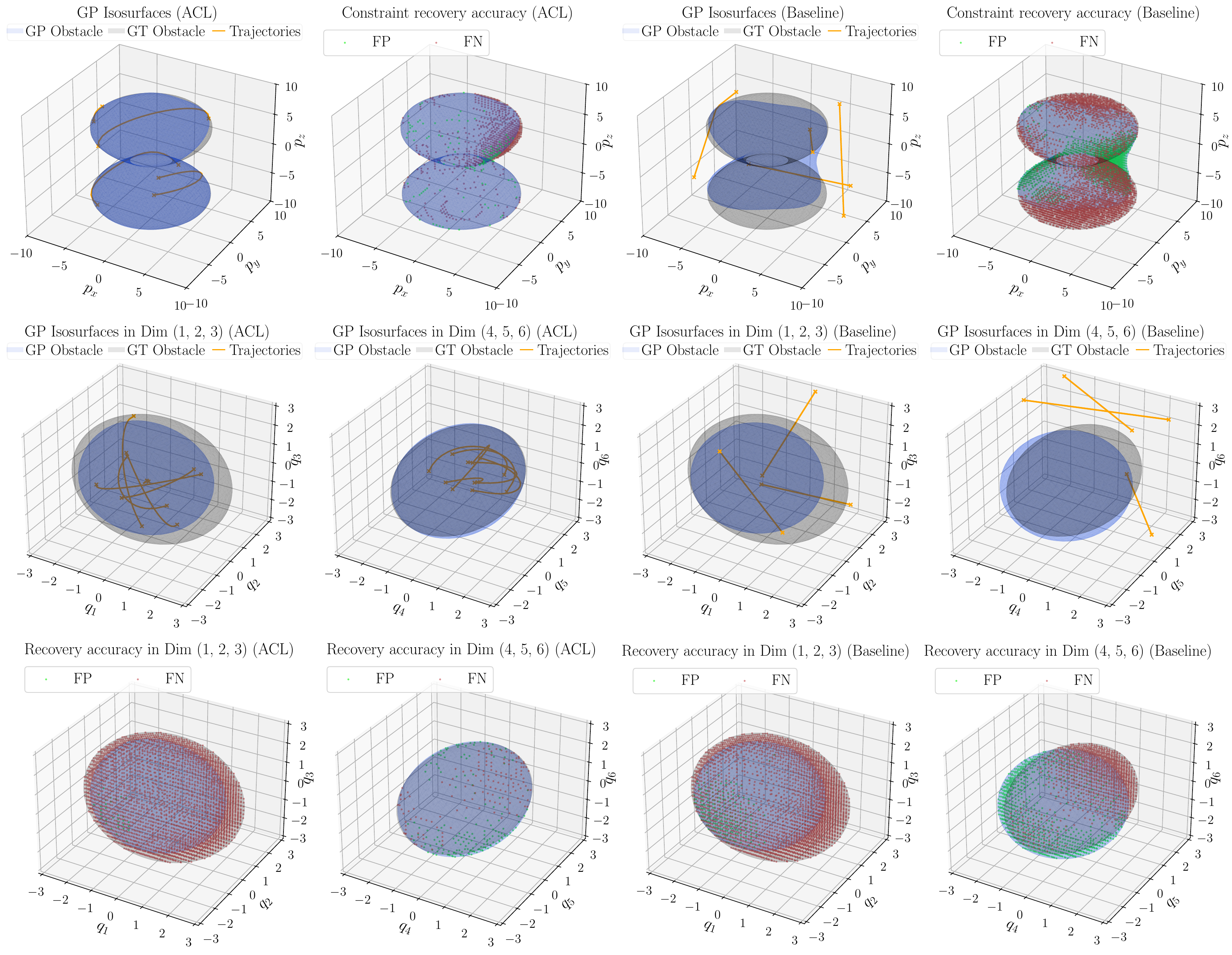}\vspace{-10pt}
    \caption{
    Our GP-ACL algorithm (left half) outperforms the random sampling baseline (right half) in accurately recovering constraints $g_{\urk,3}^\star$ (top), $g_{\urk,7}^\star$ (middle/bottom), from unicycle dynamics (top) and simulated 7-DOF arm (middle/bottom) demonstrations, with fewer false positive (green) and false negative (red) errors.
    Middle and bottom row figures display 3D slices of the 7D constraint space from our numerical simulations on the 7-DOF arm.
    }\vspace{-15pt}
    \label{fig: Exp_fig_4___quadcopter_arm}
\end{figure}

\paragraph{Quadcopter Simulations}
Our GP-ACL algorithm also outperforms the baseline method in accurately recovering constraints from demonstrations of length $T=20$ generated from high-dimensional quadcopter dynamics and the hourglass-shaped constraint $g_{\urk,5}^\star$ (Fig. \ref{fig: Exp_fig_4___quadcopter_arm}).
Here, we set $\alpha = 0.3$ when sampling GP posteriors.
Over $n_s = 27,000$ sampled constraint states, our GP-ACL algorithm achieved a higher accuracy rate ($\gamma_\ours = 0.9919$) compared to the baseline method ($\gamma_\BL = 0.9423$).
Our experiment results verify that, compared to the baseline approach, our GP-ACL algorithm achieve superior constraint recovery accuracy when learning from demonstrations generated using high-dimensional nonlinear dynamics.

\paragraph{7-DOF Arm Simulations and Hardware Experiments}
Across constraint recovery tasks involving demonstrations of length $T = 20$ generated in simulation (resp., on hardware) using 7-DOF arm dynamics and the ellipse-shaped constraint $g_{\urk,7}^\star$ (resp., using the physical obstacle visualized in Fig. \ref{fig: Front figure}), our GP-AL algorithm likewise achieves higher constraint inference accuracy compared to the baseline method.
(We do not report $\gamma_\ours$ and $\gamma_\BL$ 
here, due to challenges inherent in sampling from a 7D constraint space.) 
For 7-DOF robot arm simulations and hardware experiments, we use $\delta = 0.1$ in our GP-ACL algorithm. Moreover, for hardware experiments, we set $\alpha = 0.1$ when sampling GP posteriors.
Our experiment results illustrate that, when learning from either simulated or hardware demonstrations generated using high-dimensional robot arm dynamics, our GP-ACL algorithm achieves superior constraint recovery accuracy compared to the baseline approach.

\paragraph{Constraint Accuracy of GP-ACL vs. Baseline}

\begin{figure}[ht]
    \centering
    \includegraphics[width=0.99\linewidth]{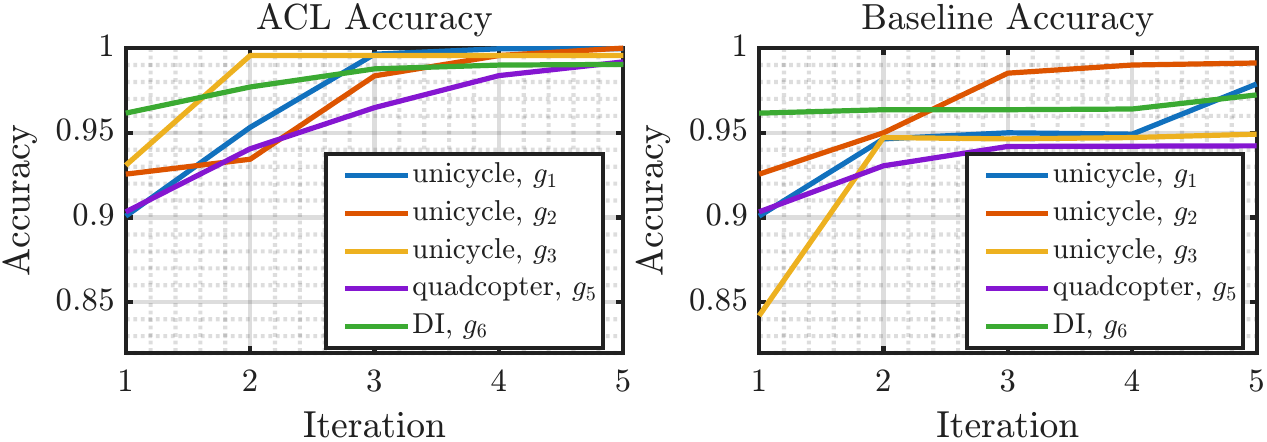}
    \caption{
    Constraint recovery accuracy of (a) our GP-ACL algorithm and (b) the random-sampling baseline approach. 
    Across the following constraint learning tasks, our GP-ACL method consistently recovered the \textit{a priori} unknown constraints more accurately compared to the random-sampling baseline method: Learning the constraints $g_{\urk,1}^\star$, $g_{\urk,2}^\star$, and $g_{\urk,3}^\star$ from demonstrations generated using unicycle dynamics (blue, red, and yellow, respectively); learning the constraint $g_{\urk,5}^\star$ from quadcopter dynamics (purple); and learning the constraint $g_{\urk,6}^\star$ from double integrator dynamics (green).
    }
    \label{fig: 0_Accuracy}
\end{figure}

In Fig. \ref{fig: 0_Accuracy} we plot constraint learning accuracy as a function of iteration count for both our GP-ACL method and the random-sampling baseline approach, 
when learning from
demonstrations generated using unicycle, quadcopter, and 3D double integrator dynamics.
Overall, our GP-ACL algorithm consistently achieves higher per-iterate constraint learning accuracy compared to the baseline sampling approach.

\section{Conclusion}
\label{sec: Conclusion}

This paper presents our Active Constraint Learning (ACL) method, which efficiently infers unknown constraints through iterative, uncertainty-guided demonstration generation. 
Across simulation and hardware experiments encompassing 
nonlinear, high-dimensional robot dynamics and non-convex, high-dimensional constraints, our ACL method successfully queries a small number of informative demonstrations to efficiently and accurately recover unknown constraints. In contrast to existing constraint inference techniques which learn from demonstrations generated without considerations of constraint uncertainty, our ACL method achieves higher constraint inference accuracy while learning from a substantially smaller, but more informative, demonstration dataset.

\newpage

\bibliography{references}

\begin{thebibliography}{28}
\providecommand{\natexlab}[1]{#1}
\providecommand{\url}[1]{\texttt{#1}}
\expandafter\ifx\csname urlstyle\endcsname\relax
  \providecommand{\doi}[1]{doi: #1}\else
  \providecommand{\doi}{doi: \begingroup \urlstyle{rm}\Url}\fi

\bibitem[Akrour et~al.(2012)Akrour, Schoenauer, and Sebag]{Akrour2012APRILActivePreferenceLearningBasedRL}
Riad Akrour, Marc Schoenauer, and Mich{\`e}le Sebag.
\newblock April: Active preference learning-based reinforcement learning.
\newblock In Peter~A. Flach, Tijl De~Bie, and Nello Cristianini, editors, \emph{Machine Learning and Knowledge Discovery in Databases}, pages 116--131, Berlin, Heidelberg, 2012. Springer Berlin Heidelberg.

\bibitem[Andersson et~al.(2019)Andersson, Gillis, Horn, Rawlings, and Diehl]{Andersson2019Casadi}
Joel A~E Andersson, Joris Gillis, Greg Horn, James~B Rawlings, and Moritz Diehl.
\newblock {CasADi: A Software Framework for Nonlinear Optimization and Optimal Control}.
\newblock \emph{Mathematical Programming Computation}, 2019.

\bibitem[Armesto et~al.(2017)Armesto, Bosga, Ivan, and Vijayakumar]{Armesto2017EfficientLearningofConstraintsandGenericNullSpacePolicies}
Leopoldo Armesto, Jorren Bosga, Vladimir Ivan, and Sethu Vijayakumar.
\newblock {Efficient Learning of Constraints and Generic Null Space Policies}.
\newblock In \emph{2017 IEEE International Conference on Robotics and Automation (ICRA)}, pages 1520--1526, 2017.

\bibitem[Chou et~al.(2020{\natexlab{a}})Chou, Ozay, and Berenson]{Chou2020LearningConstraintsfromLocallyOptimalDemonstrations}
Glen Chou, Necmiye Ozay, and Dmitry Berenson.
\newblock {Learning Constraints from Locally-optimal Demonstrations under Cost Function Uncertainty}.
\newblock \emph{IEEE Robotics and Automation Letters}, 5\penalty0 (2):\penalty0 3682--3690, 2020{\natexlab{a}}.

\bibitem[Chou et~al.(2020{\natexlab{b}})Chou, Ozay, and Berenson]{Chou2020LearningParametricConstraintsinHighDimensions}
Glen Chou, Necmiye Ozay, and Dmitry Berenson.
\newblock {Learning Parametric Constraints in High Dimensions from Demonstrations}.
\newblock In \emph{Conference on robot learning}, pages 1211--1230. PMLR, 2020{\natexlab{b}}.

\bibitem[Chou et~al.(2021)Chou, Berenson, and Ozay]{Chou2021UncertaintyAwareConstraintLearning}
Glen Chou, Dmitry Berenson, and Necmiye Ozay.
\newblock {Uncertainty-aware Constraint Learning for Adaptive Safe Motion Planning from Demonstrations}.
\newblock In \emph{Conference on robot learning}, pages 1612--1639. PMLR, 2021.

\bibitem[Chou et~al.(2022)Chou, Wang, and Berenson]{Chou2022GaussianProcessConstraintLearning}
Glen Chou, Hao Wang, and Dmitry Berenson.
\newblock {Gaussian Process Constraint Learning for Scalable Chance-constrained Motion Planning from Demonstrations}.
\newblock \emph{IEEE Robotics and Automation Letters}, 7\penalty0 (2):\penalty0 3827--3834, 2022.

\bibitem[Fang et~al.(2017)Fang, Li, and Cohn]{Fang2017LearningHowToActivelyLearn}
Meng Fang, Yuan Li, and Trevor Cohn.
\newblock {Learning how to Active Learn: A Deep Reinforcement Learning Approach}.
\newblock In Martha Palmer, Rebecca Hwa, and Sebastian Riedel, editors, \emph{{Proceedings of the 2017 Conference on Empirical Methods in Natural Language Processing}}. Association for Computational Linguistics, 2017.

\bibitem[Hu and Fisac({2023})]{HuFisac2023ActiveUncertaintyReductionforHumanRobotInteraction}
Haimin Hu and Jaime~F. Fisac.
\newblock {Active Uncertainty Reduction for Human-Robot Interaction: An Implicit Dual Control Approach}.
\newblock In \emph{{Springer}}, {Springer Proceedings in Advanced Robotics}, pages {385--401}. {Springer Nature}, {2023}.

\bibitem[Lee et~al.(2021)Lee, Smith, and Abbeel]{Lee2021PEBBLE}
Kimin Lee, Laura~M. Smith, and P.~Abbeel.
\newblock {PEBBLE: Feedback-Efficient Interactive Reinforcement Learning via Relabeling Experience and Unsupervised Pre-training}.
\newblock In \emph{International Conference on Machine Learning}, 2021.

\bibitem[Li et~al.(2025{\natexlab{a}})Li, Wang, Kar, and Jin]{LiWang2025BayesianOptimizationWithActiveConstraintLearning}
Guoyan Li, Yujia Wang, Swastik Kar, and Xiaoning Jin.
\newblock {Bayesian Optimization with Active Constraint Learning for Advanced Manufacturing Process Design}.
\newblock \emph{IISE Transactions}, 0\penalty0 (0):\penalty0 1--15, 2025{\natexlab{a}}.

\bibitem[Li et~al.(2024)Li, Siththaranjan, Sojoudi, Tomlin, and Bajcsy]{LiBajcsy2024IntentDemonstrationinGeneralSumDynamicGames}
Jingqi Li, Anand Siththaranjan, Somayeh Sojoudi, Claire Tomlin, and Andrea Bajcsy.
\newblock {Intent Demonstration in General-Sum Dynamic Games via Iterative Linear-Quadratic Approximations}.
\newblock \emph{ArXiv}, abs/2402.10182, 2024.

\bibitem[Li et~al.(2025{\natexlab{b}})Li, Chen, Yang, and Liu]{LiChen2025CooperativeActiveLearningBasedDualControlforExplorationandExploitationinAutonomousSearch}
Zhongguo Li, Wen-Hua Chen, Jun Yang, and Cunjia Liu.
\newblock {Cooperative Active Learning-Based Dual Control for Exploration and Exploitation in Autonomous Search}.
\newblock \emph{IEEE Transactions on Neural Networks and Learning Systems}, 36\penalty0 (2):\penalty0 2221--2233, 2025{\natexlab{b}}.

\bibitem[McPherson et~al.(2021)McPherson, Stocking, and Sastry]{McPherson2021MLConstraintInferenceFromStochasticDemonstrations}
David McPherson, Kaylene Stocking, and Shankar Sastry.
\newblock {Maximum Likelihood Constraint Inference from Stochastic Demonstrations}.
\newblock In \emph{CCTA}, 2021.

\bibitem[Menner et~al.(2021)Menner, Worsnop, and Zeilinger]{Menner2021ConstrainedIOC}
Marcel Menner, Peter Worsnop, and Melanie~N. Zeilinger.
\newblock {Constrained Inverse Optimal Control With Application to a Human Manipulation Task}.
\newblock \emph{IEEE TCST}, 2021.

\bibitem[Mesbah(2018)]{Mesbah2018StochasticMPCwithActiveUncertaintyLearning}
Ali Mesbah.
\newblock {Stochastic model predictive control with active uncertainty learning: A Survey on dual control}.
\newblock \emph{Annual Reviews in Control}, 45:\penalty0 107--117, 2018.
\newblock ISSN 1367-5788.

\bibitem[Murray et~al.(1994)Murray, Sastry, and Zexiang]{MurraySastry1994AMathematicalIntroductiontoRoboticManipulation}
Richard~M. Murray, S.~Shankar Sastry, and Li~Zexiang.
\newblock \emph{{A Mathematical Introduction to Robotic Manipulation}}.
\newblock CRC Press, Inc., USA, 1st edition, 1994.
\newblock ISBN 0849379814.

\bibitem[Papadimitriou and Brown(2024)]{Papadimitriou2024BayesianConstraintInferencefromUserDemonstrationsBasedonMarginRespectingPreferenceModels}
Dimitris Papadimitriou and Daniel~S. Brown.
\newblock {Bayesian Constraint Inference from User Demonstrations Based on Margin-Respecting Preference Models}.
\newblock In \emph{2024 IEEE International Conference on Robotics and Automation (ICRA)}, pages 15039--15046, 2024.

\bibitem[Papadimitriou and Li(2023)]{PapadimitriouLi2023ConstraintInferenceinControlTasksfromExpertDemonstrationsviaInverseOptimization}
Dimitris Papadimitriou and Jingqi Li.
\newblock {Constraint Inference in Control Tasks from Expert Demonstrations via Inverse Optimization}.
\newblock In \emph{2023 62nd IEEE Conference on Decision and Control (CDC)}, pages 1762--1769, 2023.

\bibitem[Papadimitriou et~al.(2022)Papadimitriou, Anwar, and Brown]{Papadimitriou2022BayesianMethodsForConstraintInferenceInRL}
Dimitris Papadimitriou, Usman Anwar, and Daniel~S. Brown.
\newblock {Bayesian Methods for Constraint Inference in Reinforcement Learning}.
\newblock \emph{Transactions on Machine Learning Research}, 2022.

\bibitem[Rasmussen and Williams(2006)]{Rasmussen2006GaussianProcessesforMachineLearning}
Carl~Edward Rasmussen and Christopher K.~I. Williams.
\newblock \emph{{Gaussian Processes for Machine Learning}}.
\newblock MIT Press, Cambridge, MA, 2006.

\bibitem[Sabatino(2015)]{sabatino2015PhDThesisquadrotor}
Francesco Sabatino.
\newblock {Quadrotor Control: Modeling, Nonlinear Control Design, and Simulation}.
\newblock Ms thesis, KTH, 2015.

\bibitem[Sadigh et~al.(2016)Sadigh, Sastry, Seshia, and Dragan]{Sadigh2016InformationGatheringActionsoverHumanInternalState}
Dorsa Sadigh, S.~Shankar Sastry, Sanjit~A. Seshia, and Anca Dragan.
\newblock {Information Gathering Actions over Human Internal State}.
\newblock In \emph{2016 IEEE/RSJ International Conference on Intelligent Robots and Systems (IROS)}, pages 66--73, 2016.

\bibitem[Sadigh et~al.(2018)Sadigh, Landolfi, Sastry, Seshia, and Dragan]{Sadigh2018PlanningforCarsthatCoordinatewithPeople}
Dorsa Sadigh, Nick Landolfi, Shankar Sastry, Sanjit Seshia, and Anca Dragan.
\newblock {Planning for Cars that Coordinate with People: Leveraging Effects on Human Actions for Planning and Active Information Gathering over Human Internal State}.
\newblock \emph{Autonomous Robots}, 42, 10 2018.

\bibitem[Singh et~al.(2018)Singh, Lacotte, Majumdar, and Pavone]{Singh2018RiskSensitiveInverseReinforcementLearning}
Sumeet Singh, Jonathan Lacotte, Anirudha Majumdar, and Marco Pavone.
\newblock {Risk-sensitive Inverse Reinforcement Learning via Semi- and Non-parametric Methods}.
\newblock \emph{The International Journal of Robotics Research}, 37\penalty0 (13-14):\penalty0 1713--1740, 2018.

\bibitem[Stocking et~al.(2022)Stocking, McPherson, Matthew, and Tomlin]{Stocking2022MaximumLikelihoodConstraintInferenceonContinuousStateSpaces}
Kaylene~C. Stocking, D.~Livingston McPherson, Robert~P. Matthew, and Claire~J. Tomlin.
\newblock {Maximum Likelihood Constraint Inference on Continuous State Spaces}.
\newblock In \emph{2022 International Conference on Robotics and Automation (ICRA)}, pages 8598--8604, 2022.

\bibitem[W{\"a}chter and Biegler(2006)]{wachter2006IPOPT}
Andreas W{\"a}chter and Lorenz~T Biegler.
\newblock {On the Implementation of an Interior-point Filter Line-search Algorithm for Large-scale Nonlinear Programming}.
\newblock \emph{Mathematical Programming}, 106:\penalty0 25--57, 2006.

\bibitem[Wilson et~al.(2020)Wilson, Borovitskiy, Terenin, Mostowsky, and Deisenroth]{wilson2020efficiently}
James Wilson, Viacheslav Borovitskiy, Alexander Terenin, Peter Mostowsky, and Marc Deisenroth.
\newblock {Efficiently Sampling Functions from Gaussian Process Posteriors}.
\newblock In \emph{International Conference on Machine Learning}, pages 10292--10302. PMLR, 2020.

\end{thebibliography}

\newpage

\appendix

\section{Supplementary Material for Sec. \ref{sec: Preliminaries and Problem Formulation}}
\label{sec: App, Supplementary Material for Preliminaries and Problem Statement}

\subsection{Extraction of Tight Points and Constraint Gradient Estimates}
\label{subsec: App, Extraction of Tight Points and Constraint Gradient Estimates}

Our work uses methods first proposed in Sec. IV-A in \cite{Chou2022GaussianProcessConstraintLearning} to extract tight system states and compute the associated gradient values. We describe these methods below for completeness.

First, we describe the extraction of \textit{tight} system states from a given locally-optimal demonstration $\xi_d$. Complementary slackness \eqref{Eqn: KKT, Forward problem, Complementary slackness} encodes that if the constraint $g_\urk(\phi_\sep(x_t(\xi_d))) \leq 0$ were \textit{not} tight at some time $t \in [T]$, the corresponding Lagrange multiplier 
at that time $t \in [T]$ must be zero. In that case, we could enforce the stationarity condition at time $t \in [T]$ while setting the unknown constraint's Lagrange multiplier $\boldlambda_{d,\urk}$ equal to $\boldzero$, i.e., $\bolds_{x_t}(\xi_d, \boldlambda_{d,k}, \boldzero, \boldnu_d) = \boldzero$. Conversely, if $g_\urk(\phi_\sep(x_t(\xi_d))) = 0$, then $\bolds_{x_t}(\xi_d, \boldlambda_{d,k}, \boldzero, \boldnu_d) \ne \boldzero$ for all possible $\boldlambda_{d,k}$ and $\boldnu_d$ satisfying dual feasibility \eqref{Eqn: KKT, Forward problem, Dual feasibility} and complementary slackness \eqref{Eqn: KKT, Forward problem, Complementary slackness}, and thus the following linear program would yield a strictly positive optimal value (\cite{Chou2022GaussianProcessConstraintLearning}, Corollary 1):

\begin{problem}[Tightness Check] \label{Prob: Tightness check}
\begin{subequations}
\begin{align}
	\min. \hspace{5mm} &\Vert \emph{\bolds}_{x_t}(\xi_d, \boldlambda_{d,k}, \emph{\boldzero}, \boldnu_d)  \Vert_1 \\
	\emph{\st} \hspace{5mm} &\boldlambda_{d,k} \geq \emph{\boldzero}, \hspace{5mm} \boldlambda_{d,k} \odot \emph{\boldg}_k(\xi_d) = \emph{\boldzero}.
\end{align}
\end{subequations}
\end{problem}

For each $d \in [D]$, we collect times $t \in [T]$ at which Prob. \ref{Prob: Tightness check} yields a strictly positive optimal value to construct an \textit{estimated set of tight timesteps} $\boldt_\tight(\xi_d)$, which is a guaranteed under-approximation of the true set of tight timesteps $\{t \in [T]: g_\urk^\star(\phi_\sep(x_t(\xi_d))) = 0 \}$ for the $d$-th demonstration.

Next, for each demonstration $d \in [D]$, at each 
$t \in \boldt_\tight(\xi_d)$, we describe a method for generating estimates $w_{d,t}$ for the corresponding gradient of the unknown constraint, i.e., $\nabla_{x_t} g_\urk^\star(\phi_\sep(x_t(\xi_d)))$. Concretely, following Sec. IV-A in \cite{Chou2022GaussianProcessConstraintLearning}, we first define $\boldone_\tight(\xi_d) \in \R^T$ to be the vector whose $t$-th component equals 1 if $t \in \boldt_\tight(\xi_d)$, and 0 otherwise, for each $t \in [T]$. Then, we fix $\boldlambda_{d, \urk} = \boldone_\tight(\xi_d)$, and solve for gradient estimates $w_{d,t} \in \R^{1 \times n}$ which are compatible with the KKT conditions \eqref{Eqn: KKT, Forward problem} via the following linear program:

\begin{problem}[Constraint Gradient Estimation] \label{Prob: Constraint Gradient Estimation}
\begin{subequations}
\begin{align}
	\emph{\find}. \hspace{5mm} &\boldlambda_{d,k}, \boldnu_d, \{w_{d,t}: t \in \emph{\boldt}_\emph{\tight}(\xi_d) \} \\
	\emph{\st} \hspace{5mm} &\emph{\boldg}_\urk^\star(\phi(\xi_d)) \leq \emph{\boldzero}, \hspace{5mm} \boldlambda_{d,k} \geq \emph{\boldzero}, \hspace{5mm} \boldlambda_{d,k} \odot \emph{\boldg}_k(\xi_d) = \emph{\boldzero}, \\
	&\nabla_{x_t} c(\xi) + \boldlambda_k^\top \nabla_{x_t} \emph{\boldg}_k^\star(\xi) + \mathbbm{1}\{t \in \emph{\boldt}_\emph{\tight}(\xi_d) \} \cdot w_{d,t} + \boldnu_k^\top \nabla_{x_t} \emph{\boldh}_k(\xi) = \emph{\boldzero}, \ \forall \ t \in [T], \\
    &\nabla_{u_t} c(\xi) + \boldlambda_k^\top \nabla_{u_t} \emph{\boldg}_k^\star(\xi) + \boldnu_k^\top \nabla_{u_t} \emph{\boldh}_k(\xi) = \emph{\boldzero}, \ \forall \ t \in [T].
\end{align}
\end{subequations}
\end{problem}

Thm. 2 in \cite{Chou2022GaussianProcessConstraintLearning} provides mild 
conditions 
under which 
$\{w_{d,t}: t \in \boldt_\tight(\xi_d)\}$ recover the true constraint gradients $\{\nabla_{x_t} g_\urk^\star\big(\phi_\sep(x_t(\xi_d)) \big): t \in \boldt_\tight(\xi_d) \}$ up to scale, i.e., for each $t \in \boldt_\tight(\xi_d)$, there exists some $\alpha_{d,t} > 0$ such that $\nabla_{x_t} g_\urk^\star\big(\phi_\sep(x_t(\xi_d)) \big) = \alpha_{d,t} w_{d,t}$.

\subsection{A Primer on Gaussian Processes and Their Training Process}
\label{subsec: App, A Primer on Gaussian Processes and Their Training Process}

A Gaussian Process $\GP(m, k)$ is a (possibly infinite) set of random variables characterized by a mean function $m: \R^n \ra \R$ and a covariance function $k: \R^n \times \R^n \ra \R$, whose finite sub-collections are always jointly Gaussian (\cite{Rasmussen2006GaussianProcessesforMachineLearning}). GPs are often employed as priors in regression tasks, in which one aims to infer an unknown map $g: \R^n \ra \R$ from an input-output dataset $\D := (\{(x_i, y_i) \in \R^n \times \R^m \})_{i=1}^{N_d}$ generated via a noisy output model $y_i \sim \mathcal{N}(g(x_i), \sigma^2)$. In such scenarios, the posterior $(\tilde g|\D)(\cdot)$ of the function $g$ is characterized by the following posterior mean and covariance maps:
\begin{subequations} \label{Eqn: GP, Posterior Mean and Covariance}
\begin{align} \label{Eqn: GP, Posterior Mean}
	\E[\tilde g(\cdot)|\D] &= k(\cdot, \boldX) \big( k(\boldX, \boldX) + \sigma^2 I \big)^{-1} \boldY, \\ \label{Eqn: GP, Posterior Covariance}
	\cov[ \tilde g(\cdot)|\D] &= k(\cdot, \cdot) - k(\cdot, \boldX) \big( k(\boldX, \boldX) + \sigma^2 I \big)^{-1} k(\boldX, \cdot)
\end{align}
\end{subequations}

Whereas \cite{Chou2022GaussianProcessConstraintLearning} only uses the posterior mean for constraint inference, we additionally sample from the posterior $(\tilde g|\D)(\cdot)$ to facilitate higher constraint learning efficiency.
Concretely, we sample posterior functions $\{\hat g_p(\cdot): p \in [P] \}$ using the Random Fourier Features-based approach presented in \cite{wilson2020efficiently}. Concretely, we sample $n_\ell$ basis functions $\phi_\ell: \R^{n_c} \ra \R$, as described in \cite{wilson2020efficiently}, and select a scale coefficient $\alpha > 0$, which modulates the level of random deviations between $\E[\tilde g(\cdot)|\D]$ and each posterior function $\hat g_p(\cdot)$. We then draw $w_{p, \ell} \sim \iid \ \mathcal{N}(0, 1)$, and define, for each $p \in P$:
\begin{align} \label{Eqn: GP Posterior Sampling}
    \hat g_p(\cdot) &:= \E[\tilde g(\cdot)|\D] + \alpha \cdot \sum_{\ell=1}^{n_\ell} w_{p,\ell} \Big( \phi_\ell(\cdot) -  k(\cdot, \boldX)  \big( k(\boldX, \boldX) + \sigma^2 \big)^{-1} \phi_\ell(\boldX) \Big),
\end{align}
to form the desired GP posterior samples.

\section{Supplementary Material for Sec. \ref{sec: Experiments}}
\label{sec: App, Supplementary Material for Experiments Section}

\subsection{Constraint Types}
\label{subsec: App, Constraint Types}

We concretely formulate the constraints $g_{\urk,1}^\star, \cdots, g_{\urk,7}^\star$ and associated avoid sets $\Avoid_1, \cdots, \Avoid_7$ below. (Note that $\Avoid_8$ is a physical obstacle in a hardware experiment). Recall that, by definition, $\Avoid_i := \{\kappa \in \R^{n_c}: g_{\urk, i}^\star(\kappa) \leq 0 \}$ for each $i \in [8]$. As formulated in Sec. \ref{sec: Experiments}, given a state $x$, we refer to the position, $x$-coordinate position, $y$-coordinate position, and $z$-coordinate position by $p$, $p_x$, $p_y$, and $p_z$, respectively.

For $g_{\urk,1}^\star$ and $\Avoid_1$, we set:
\begin{align}
	g_{\urk,1}^\star(\phi_\sep(x)) &= -0.02 | 1.732 p_x + p_y|^{1.4} - 0.042 |p_x - 1.732 p_y|^{1.4} + 1
\end{align}

For $g_{\urk,2}^\star$ and $\Avoid_2$, we set:
\begin{align}
    g_{\urk,2}^\star(\phi_\sep(x)) &= - \min_{i \in [2]}\{p^\top Q_i p + q_i^\top p + r_i \},
\end{align}
where:
\begin{subequations}
\begin{alignat}{3}
    Q_1 &= \begin{bmatrix}
    0.0868 & 0.0243 \\
    0.0243 & 0.0868 \\
    \end{bmatrix}, \hspace{5mm} &&q_1 = \begin{bmatrix}
        -0.8403 \\ -0.7153
    \end{bmatrix}, \hspace{5mm} &&R_1 = 1.7534 \\
    Q_2 &= \begin{bmatrix}
    0.0868 & 0.0243 \\
    0.0243 & 0.0868 \\
    \end{bmatrix}, \hspace{5mm} &&q_2 = \begin{bmatrix}
        0.8403 \\ 0.7153
    \end{bmatrix}, \hspace{5mm} &&R_2 = 1.7534.
\end{alignat}
\end{subequations}

For $g_{\urk,3}^\star$ and $\Avoid_3$, we set:
\begin{align}
    g_{\urk,3}^\star(\phi_\sep(x)) := - \left( \frac{1}{4}p_x^2 + p_y^2 - 25 \right)^2 - 50 p_x^2 + 687.5 
\end{align}

For $g_{\urk,4}^\star$ and $\Avoid_4$, we set:
\begin{align}
    g_{\urk,4}^\star(\phi_\sep(x)) := - \max\left\{ \frac{1}{5}|p_x|, \frac{1}{4}|p_y| \right\} + 1.
\end{align}

For $g_{\urk,5}^\star$ and $\Avoid_5$, we set:
\begin{align}
    g_{\urk,5}^\star(\phi_\sep(x)) := -\left(\frac{1}{9}x^2 + \frac{1}{9}y^2 + z^2 - 6 \right)^2 - \frac{16}{3}(x^2 + y^2) + 39.6.
\end{align}

For $g_{\urk,6}^\star$ and $\Avoid_6$, we set:
\begin{align}
    g_{\urk,6}^\star(\phi_\sep(x)) := - \min_{i \in [3]} (x - v_i)^\top W_i (x - v_i) + R,
\end{align}
where:
\begin{alignat}{2}
    v_1 &= \begin{bmatrix}
        0 \\ 0 \\ 0
    \end{bmatrix}, \hspace{5mm} &&W_1 = \begin{bmatrix}
        2 & 0 & 0 \\
        0 & 2 & 0 \\
        0 & 0 & 1
    \end{bmatrix}, \hspace{5mm} R = 49, \\
    v_2 &= \begin{bmatrix}
        5 \\ 0 \\ 5
    \end{bmatrix}, \hspace{5mm} &&W_2 = \begin{bmatrix}
        1 & 0 & 0 \\
        0 & 4 & 0 \\
        0 & 0 & 4
    \end{bmatrix}, \\
    v_3 &= \begin{bmatrix}
        -5 \\ 0 \\ -5
    \end{bmatrix}, \hspace{5mm} &&W_3 = \begin{bmatrix}
        1 & 0 & 0 \\
        0 & 4 & 0 \\
        0 & 0 & 4
    \end{bmatrix}, \\
\end{alignat}

For $g_{\urk,7}^\star$ and $\Avoid_7$, we set:
\begin{align}
    g_{\urk,7}^\star(\phi_\sep(x)) := - x^2 B^{-2} x + 1,
\end{align}
where $B := \text{diag}\{150, 90, 150, 90, 150, 90, 150\} \in \R^{7 \times 7}$. Here, $\text{diag}\{\cdot\}$ describes a diagonal matrix, with the given entries indicating diagonal values.

\subsection{Additional Experiment Results}
\label{subsec: App, Additional Experiment Resultss}

\paragraph{2D Double Integrator Experiments}

\begin{figure}[ht]
    \centering
    \includegraphics[width=0.99\linewidth]{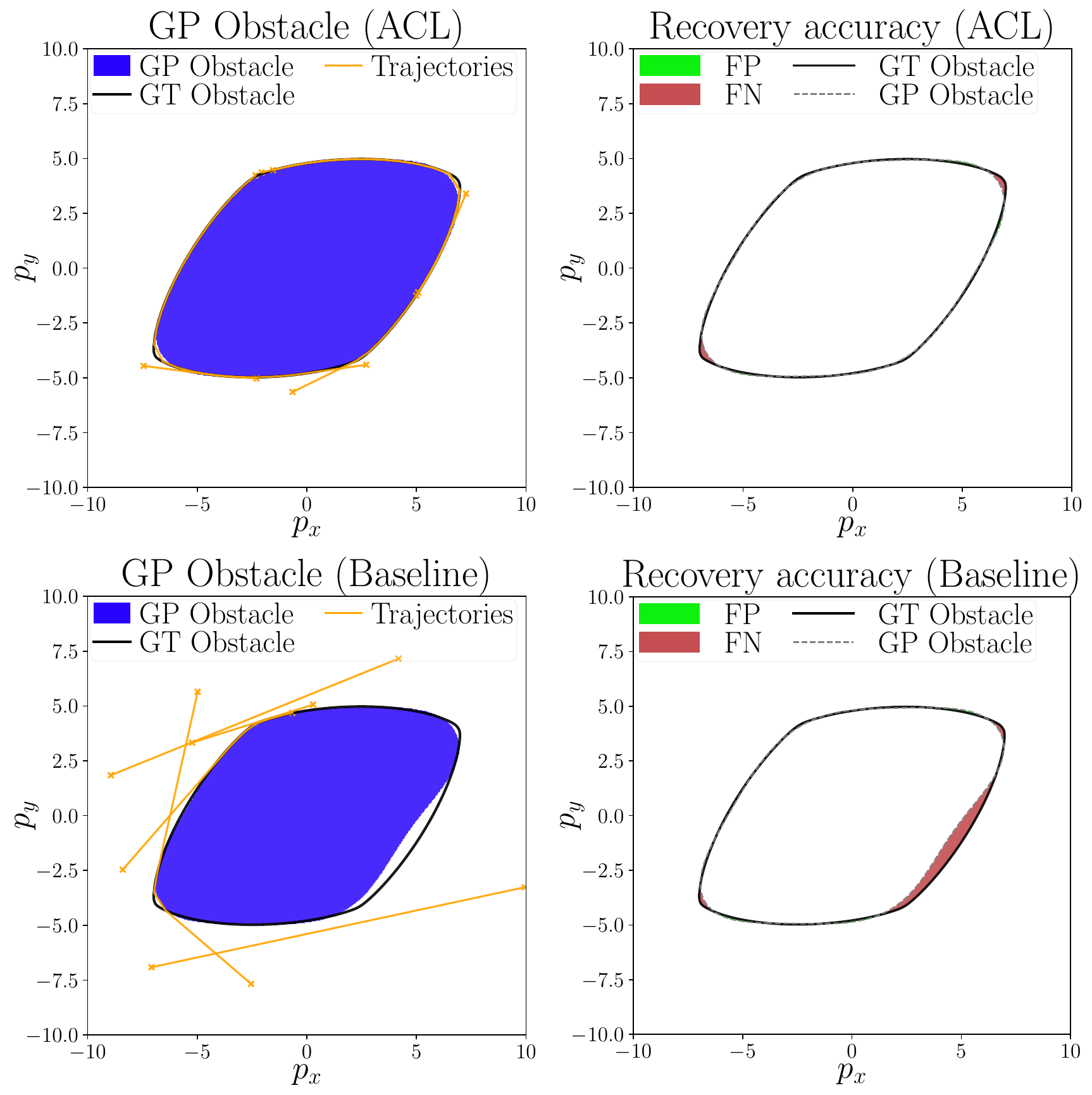}
    \caption{
    Our GP-ACL algorithm (top) outperforms the random sampling baseline (bottom) in accurately recovering the nonlinear constraint $g_{\urk,1}^\star$ from demonstrations generated using 2D double integrator dynamics, with fewer false positive (green) and fewer false negative (red) errors.
    }
    \label{fig: 5_DI_super_ellipse}
\end{figure}

\begin{figure}[ht]
    \centering
    \includegraphics[width=0.99\linewidth]{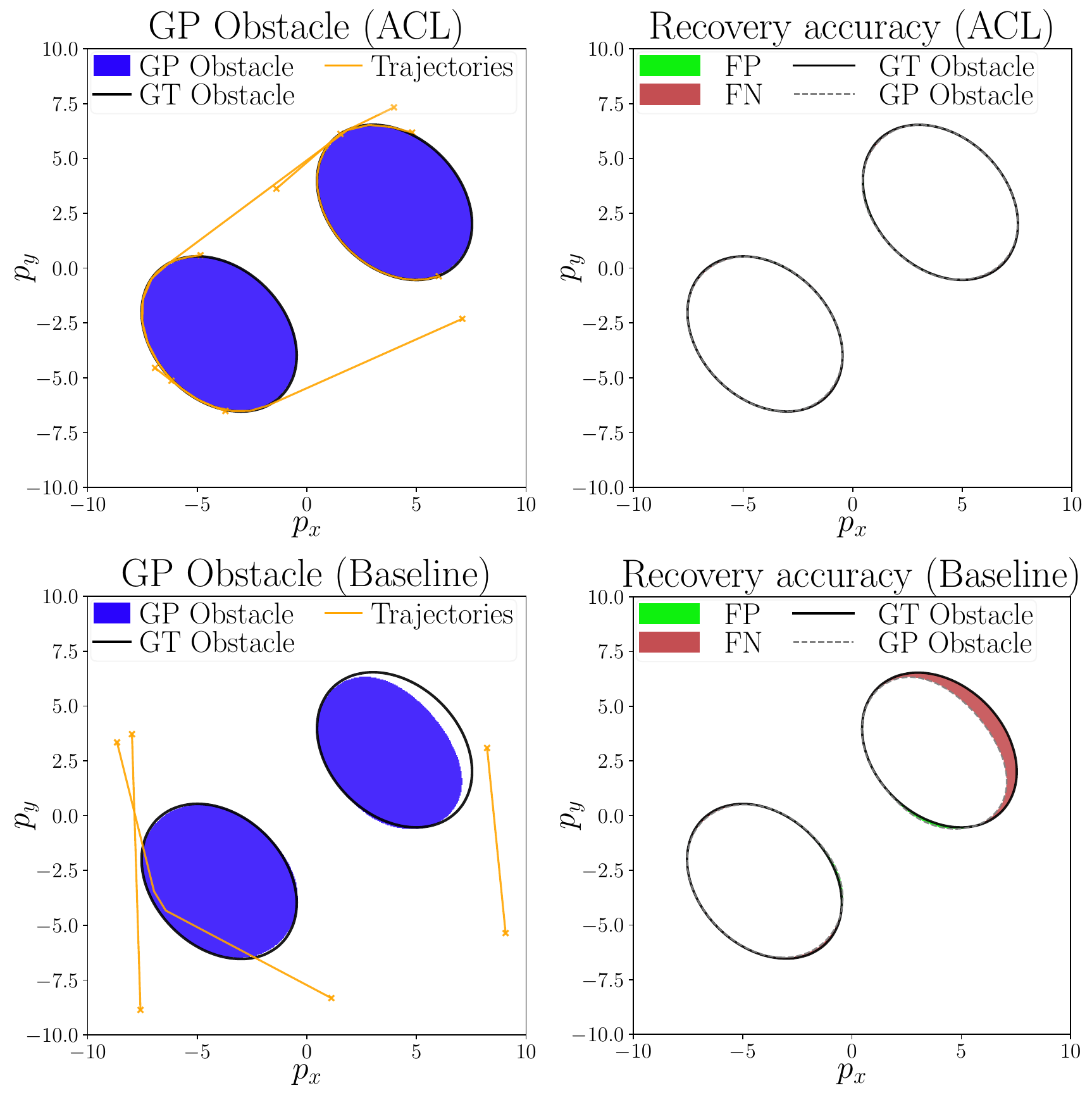}
    \caption{
    Our GP-ACL algorithm (top) outperforms the random sampling baseline (bottom) in accurately recovering the nonlinear constraint $g_{\urk,2}^\star$ from demonstrations generated using 2D double integrator dynamics, with fewer false positive (green) and fewer false negative (red) errors.
    }
    \label{fig: 6_DI_double_quadratic}
\end{figure}

\begin{figure}[ht]
    \centering
    \includegraphics[width=0.99\linewidth]{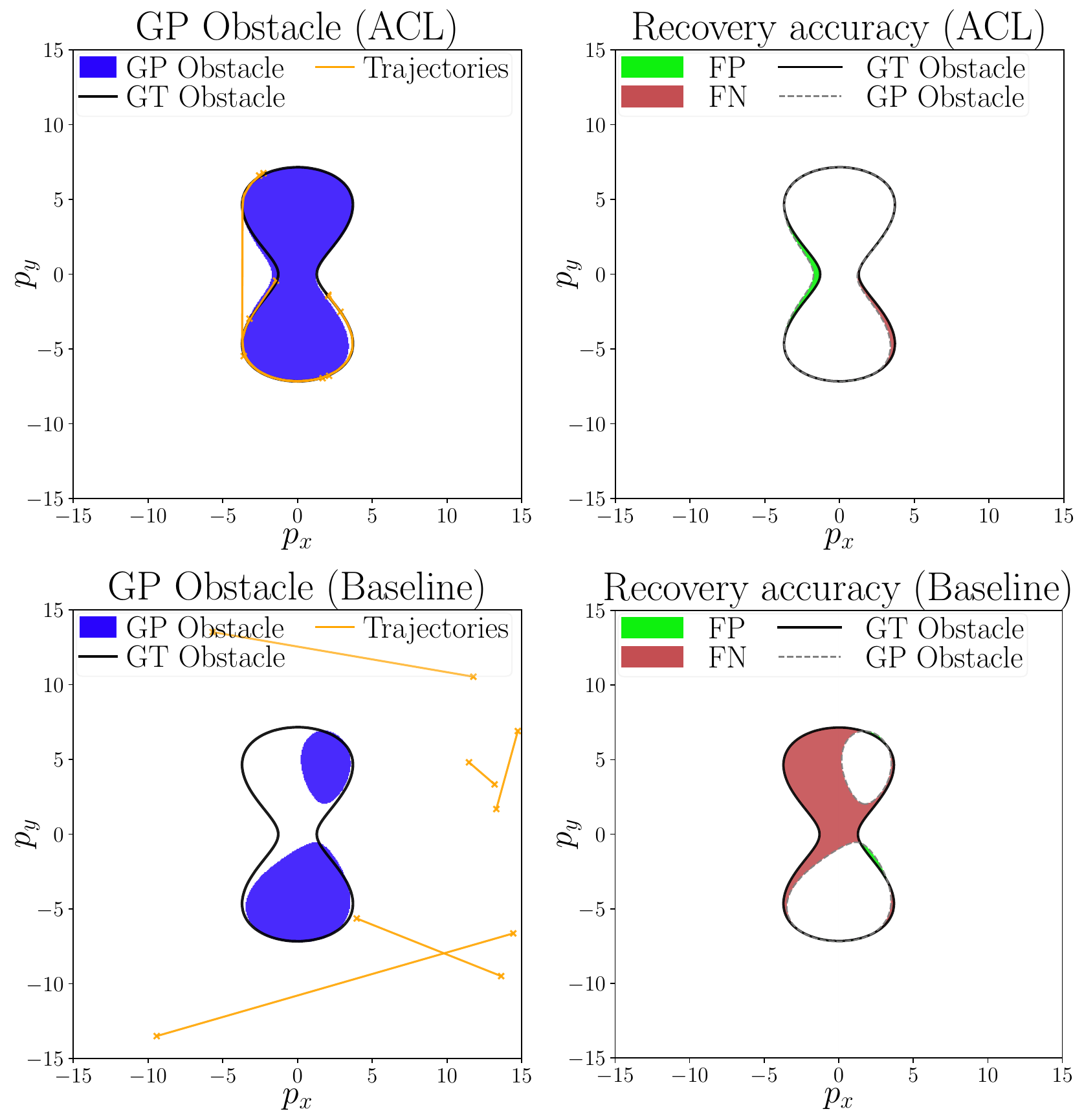}
    \caption{
    Our GP-ACL algorithm (top) outperforms the random sampling baseline (bottom) in accurately recovering the nonlinear constraint $g_{\urk,3}^\star$ from demonstrations generated using 2D double integrator dynamics, with fewer false positive (green) and fewer false negative (red) errors.
    }
    \label{fig: 7_DI_peanut}
\end{figure}

\begin{figure}[ht]
    \centering
    \includegraphics[width=0.99\linewidth]{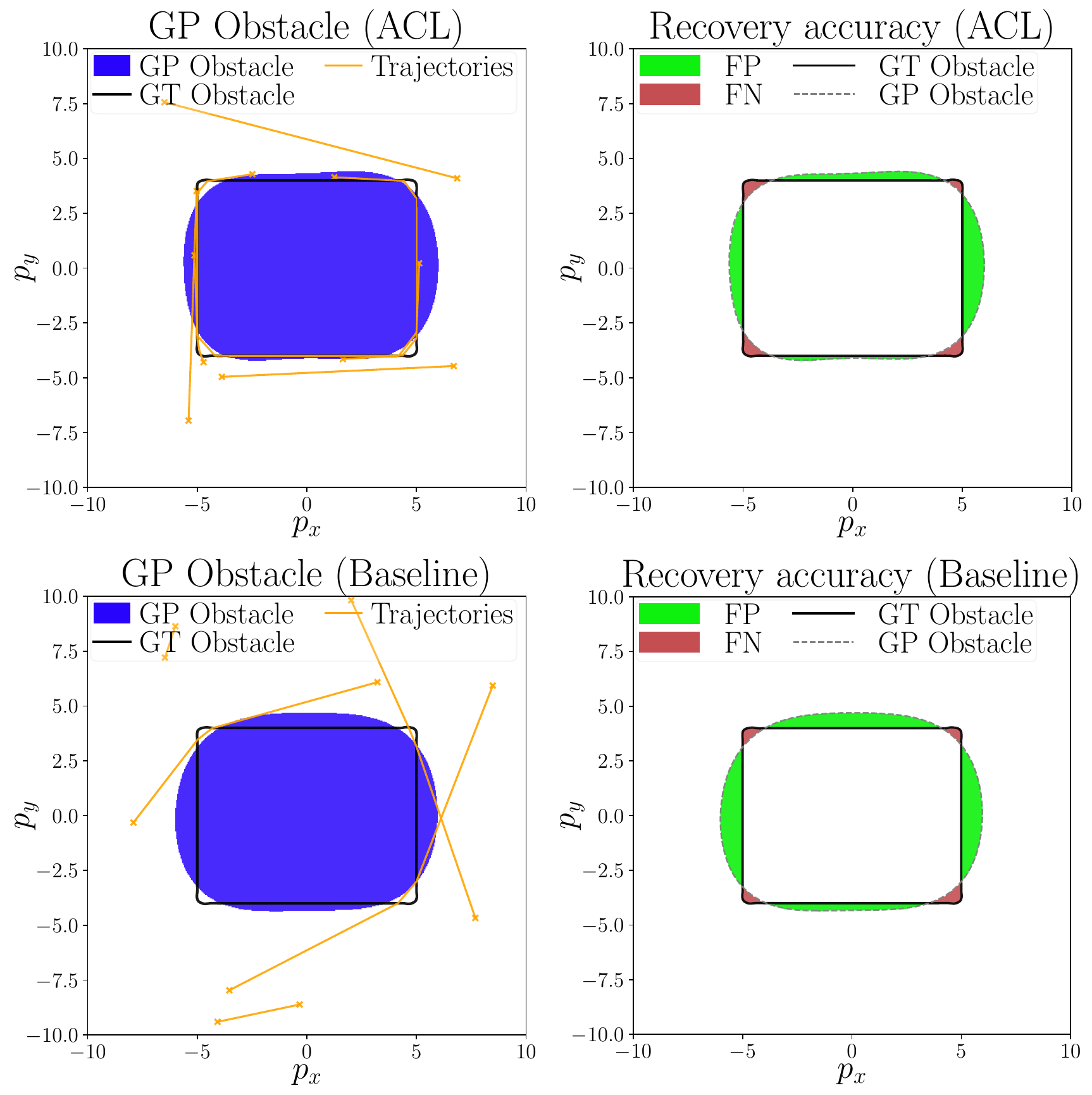}
    \caption{
    Our GP-ACL algorithm (top) outperforms the random sampling baseline (bottom) in accurately recovering the nonlinear constraint $g_{\urk,4}^\star$ from demonstrations generated using 2D double integrator dynamics, with fewer false positive (green) and fewer false negative (red) errors.
    }
    \label{fig: 8_DI_box}
\end{figure}

We evaluate our GP-ACL algorithm on the problem of recovering the three nonlinear constraints $g_{\urk,1}^\star$, $g_{\urk,2}^\star$, $g_{\urk,3}^\star$, and $g_{\urk,4}^\star$, as defined in \ref{subsec: App, Constraint Types} and visualized in Figs. \ref{fig: 5_DI_super_ellipse}, \ref{fig: 6_DI_double_quadratic}, \ref{fig: 7_DI_peanut}, and \ref{fig: 8_DI_box}.
Here, we set $\alpha = 0.3$ when sampling GP posteriors for learning $g_{\urk,3}^\star$, and use the default setting of $\alpha = 0.15$ while learning $g_{\urk,1}^\star$, $g_{\urk,2}^\star$, $g_{\urk,3}^\star$, and $g_{\urk,4}^\star$. All other parameters are set at the default values provided in Sec. \ref{subsec: Experiment Setup}.
Across $n_s = 2500$ sampled constraint states, our GP-ACL algorithm accurately predicts the safeness or unsafeness of each sample with accuracy $\gamma_\ours = 0.9952$, $0.996$, $0.9956$, and $0.9762$ for the constraints $g_{\urk,1}^\star$, $g_{\urk,2}^\star$, $g_{\urk,3}^\star$, and $g_{\urk,4}^\star$, respectively, while the random sampling-based baseline method yielded accuracies of only $\gamma_\BL = 0.99$, $0.9868$, $0.9652$, and $0.9692$, respectively.

\paragraph{3D Double Integrator Experiments}

\begin{figure}[ht]
    \centering
    \includegraphics[width=0.96\linewidth]{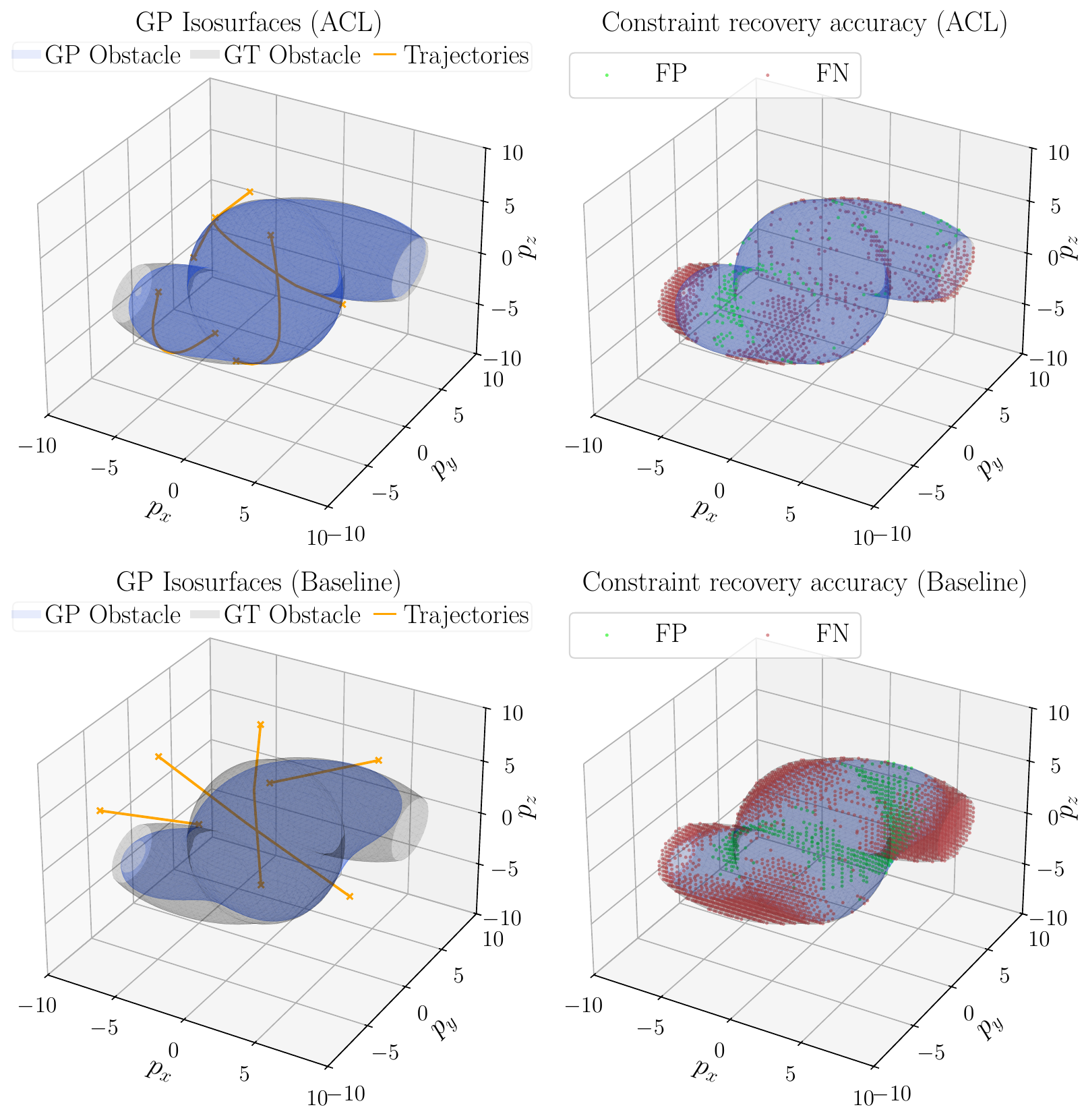}
    \caption{
    Our GP-ACL algorithm (top) outperforms the random sampling baseline (bottom) in accurately recovering the nonlinear constraint $g_{\urk,6}^\star$ from demonstrations generated using 3D double integrator dynamics, with fewer false positive (green) and false negative (red) errors.
    }
    \label{fig: DI_Z_shape}
\end{figure}

We further evaluate our GP-ACL algorithm by recovering the complex nonlinear constraint $g_{\urk,6}^\star$, as defined in \ref{subsec: App, Constraint Types} and visualized in Fig. \ref{fig: DI_Z_shape}, using demonstrations of length $T = 20$.
Across $n_s = 2500$ sampled constraint states, our GP-ACL algorithm accurately predicts the safeness or unsafeness of each sample with accuracy $\gamma_\ours = 0.9914$, while 
the random sampling-based baseline method yielded accuracies of only $\gamma_\BL = 0.9724$.

\paragraph{Unicycle Experiments (Additional)}

\begin{figure}[ht]
    \centering
    \includegraphics[width=0.96\linewidth]{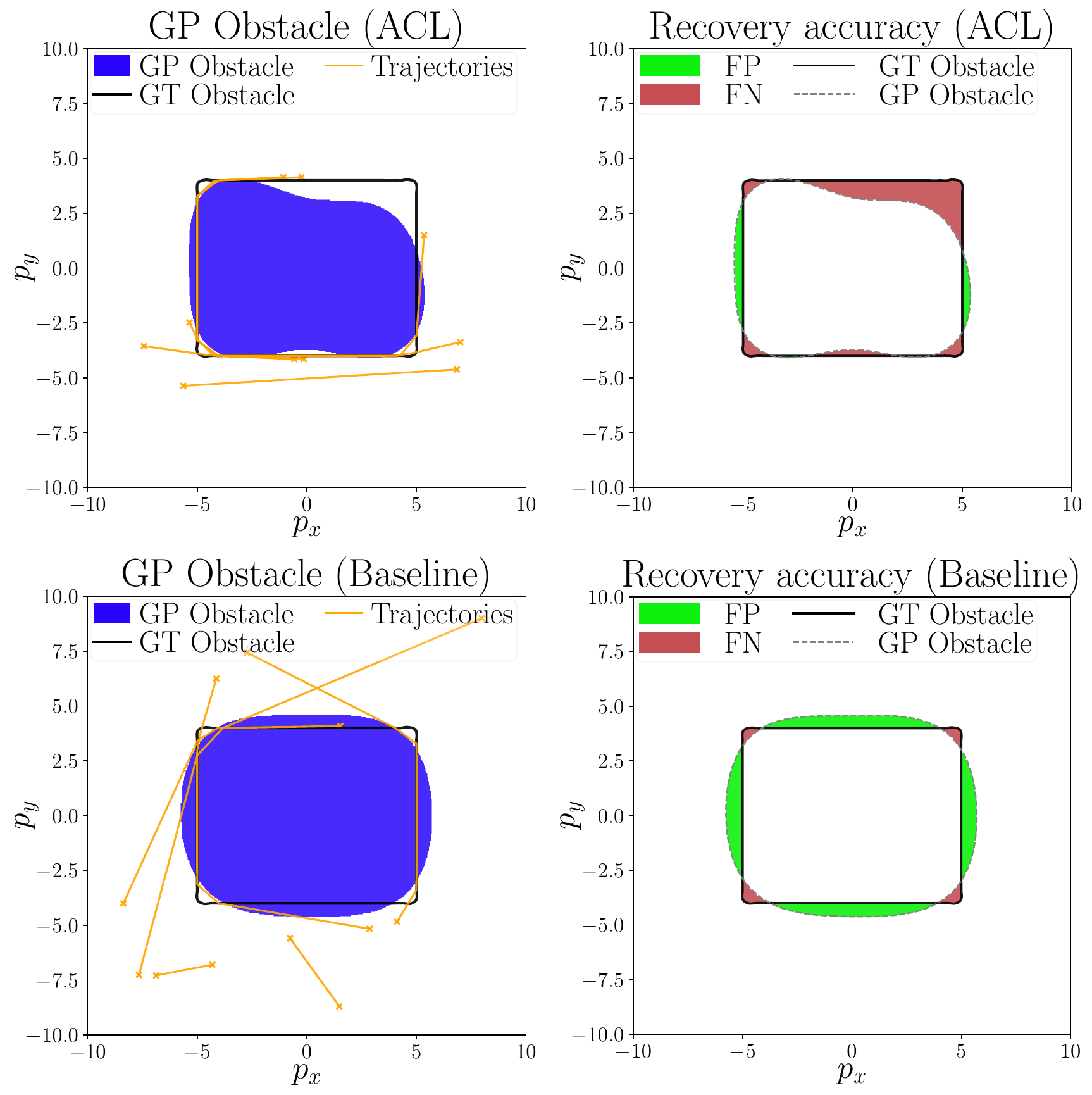}
    \caption{
    Our GP-ACL algorithm (top) outperforms the random sampling baseline (bottom) in accurately recovering the constraint $g_{\urk,4}^\star$ from demonstrations generated using 4D unicycle dynamics, with fewer false positive (green) and false negative (red) errors.
    }
    \label{fig: 12_unicycle_box}
\end{figure}

We further evaluate our GP-ACL algorithm by recovering the constraint $g_{\urk,4}^\star$, as defined in \ref{subsec: App, Constraint Types} and visualized in Fig. \ref{fig: 12_unicycle_box}, using demonstrations of length $T = 30$.
As shown in Fig. \ref{fig: 12_unicycle_box}, our GP-ACL algorithm predicts the safeness or unsafeness of each sample more accurately than the baseline method.

\end{document}